\documentclass{article} 
\usepackage{iclr2026_conference,times}

\usepackage{amsmath,amsfonts,bm}

\def\eqref#1{equation~\ref{#1}}

\def\1{\bm{1}}

\def\vd{{\bm{d}}}

\def\vl{{\bm{l}}}
\def\vm{{\bm{m}}}
\def\vn{{\bm{n}}}

\def\vw{{\bm{w}}}

\def\mA{{\bm{A}}}

\def\mC{{\bm{C}}}

\def\mE{{\bm{E}}}

\def\mH{{\bm{H}}}

\def\mK{{\bm{K}}}

\def\mO{{\bm{O}}}
\def\mP{{\bm{P}}}
\def\mQ{{\bm{Q}}}

\def\mS{{\bm{S}}}

\def\mV{{\bm{V}}}
\def\mW{{\bm{W}}}
\def\mX{{\bm{X}}}
\def\mY{{\bm{Y}}}
\def\mZ{{\bm{Z}}}

\DeclareMathAlphabet{\mathsfit}{\encodingdefault}{\sfdefault}{m}{sl}
\SetMathAlphabet{\mathsfit}{bold}{\encodingdefault}{\sfdefault}{bx}{n}

\newcommand{\R}{\mathbb{R}}

\usepackage{hyperref}
\usepackage{url}
\usepackage{graphicx}
\usepackage{nicefrac}
\usepackage{xcolor}
\usepackage{multirow}
\usepackage[table,xcdraw]{xcolor}
\usepackage{arydshln}
\usepackage{algorithm}
\usepackage{algpseudocode}
\usepackage{amsmath}
\usepackage{booktabs}
\usepackage{capt-of}   
\usepackage{wrapfig}
\usepackage{subcaption}

\title{KVLinC: KV Cache Quantization with Hadamard Rotation and Linear Correction}

\iclrfinalcopy
\author{Utkarsh Saxena, Kaushik Roy \\
Departmepnt of Electrical and Computer Engineering\\
Purdue University\\
West Lafayette, Indiana, USA. \\
\texttt{\{saxenau,kaushik\}@purdue.edu} \\
}

\begin{document}

\maketitle

\begin{abstract}
Quantizing the key-value (KV) cache is a promising strategy for improving the inference efficiency of large language models (LLMs). However, aggressive quantization to very low precision (e.g., 2 bits) introduces significant errors in the stored key and value tensors, which propagate through the dot-product attention mechanism and ultimately degrade generation quality. To address this, we propose \emph{KVLinC}, a framework to mitigate attention errors introduced by KV cache quantization in the extreme low-precision regime. KVLinC combines a Hadamard rotation, which reduces quantization error in values, with lightweight linear correction adapters that explicitly compensate for errors introduced by quantized keys. Across extensive evaluations on the LLaMA, Qwen2.5, and Qwen3 model families, KVLinC consistently matches or surpasses strong baselines while achieving higher KV-cache compression. Furthermore, we implement a custom attention kernel that results in upto $2.55\times$ faster inference compared to Flash Attention baseline, enabling efficient long-context LLM inference. Code is available at \url{https://github.com/UtkarshSaxena1/kvlinc}.
\end{abstract}

\section{Introduction}

Large Language Models (LLMs) \citep{llama3.2meta, llama3meta, yang2024qwen2, yang2025qwen3technicalreport} have achieved strong performance across diverse NLP tasks, but their deployment remains costly due to heavy memory and compute demands during inference. A major bottleneck is the key-value (KV) cache, which stores past activations in every transformer layer to enable autoregressive decoding. Unlike model parameters, which are fixed in size, the KV cache grows linearly with sequence length and batch size, quickly dominating GPU memory and bandwidth. For example, in Llama-3-8B \citep{llama3meta} with a sequence length of 8k and a batch size of 16, the KV cache alone consumes 16 GB, which is comparable to the parameter footprint. As applications push toward longer contexts or larger batch sizes, the KV cache quickly dominates memory and bandwidth requirements, limiting throughput and inflating serving costs. Thus, reducing KV cache size while preserving accuracy is critical for scaling LLMs to long-context and high-throughput settings.

Quantization of KV cache is a promising direction to reduce inference memory cost by representing the key value tensors in lower precision formats \citep{hooper2024kvquant, liu2024kivi, ashkboos2024quarot}. Recent work KIVI \citep{liu2024kivi} has demonstrated the feasibility of compressing the KV cache to as few as 2-bits per entry. However, quantizing the KV cache to low precision introduces quantization errors in the stored key and value tensors which propagate into the dot-product attention mechanism and ultimately impair language generation ability. As sequence length of a task increases, quantization errors accumulate across the stored key and value tokens, leading to compounding distortions in attention distributions. Since each decoding step reuses the corrupted representations, performance degradation becomes more severe with increasing sequence length \cite{kang2024gearefficientkvcache}.

QuaRot \citep{ashkboos2024quarot} demonstrated that applying a rotation prior to quantization can substantially reduce quantization error compared to directly quantizing the raw tensor. Specifically, QuaRot leverages Hadamard rotations to rotate the key and value tensors into a representation more suitable for low-precision storage. While this approach has shown effectiveness at moderate precision levels, such as a 4-bit KV cache, its applicability under more aggressive quantization settings remains unexplored. In contrast, another line of work focuses on compensating for quantization error by preserving selected components of the KV cache in higher precision. For example, ResQ \citep{saxena2024resq} retains critical channels in high precision, while Gear \citep{kang2024gearefficientkvcache} maintains a low-rank reconstruction of the quantization error. However, in both cases, the memory cost of storing high-precision components grows proportionally with the KV cache. At long context lengths, this overhead becomes non-negligible, limiting the overall compression benefits of KV cache quantization. 

To address this, we propose KVLinC, a framework explicitly designed to mitigate attention errors introduced by KV cache quantization in the extreme low-precision regime. KVLinC combines complementary strategies for keys and values that enable robust compression of the KV cache to 2-bit while maintaining strong performance across both short and long context tasks. First, we revisit rotation-based quantization methods and analyze their robustness at 2-bit precision. We explore different quantization axes — specifically, applying quantization along the channel axis or the token axis when combined with Hadamard rotated keys and values. Our experiments reveal that optimal performance is achieved by quantizing raw keys along the channel axis, while rotated values perform best when quantized along the token axis. 

Second, to further mitigate the impact of quantization error, we introduce linear correction adapters, trainable modules that explicitly learn to track and compensate for distortions in the attention distribution caused by quantized keys. These adapters incur only a constant memory overhead that does not grow with sequence length. Moreover, their computational cost is linear with sequence length, in contrast to quadratic complexity of self-attention, making them both efficient and practical for long-context inference. Our design is motivated by linear attention methods~\citep{zhang2024lolcats, lan2025liger}, which discard most tokens and train adapters to recover the resulting error. While effective for short contexts, such methods replace softmax with a lossy linear approximation, leading to distortions that cannot be fully corrected. In contrast, our approach retains the full token history and corrects only quantization-induced errors in keys which makes it an easier learning problem. This allows KVLinC to achieve effective compression while preserving the fidelity of softmax attention, naturally scaling to long contexts. In summary, our contributions are as follows:

\begin{itemize}
    \item We analyze the various design choices related to Hadamard rotation based KV cache quantization and observe that quantizing keys along the channel axis and quantizing Hadamard rotated values along the token axis is optimal. 
    \item We introduce linear correction adapters which are trained to correct attention error introduced by quantized KV cache. 
    \item We evaluate KVLinC on various short and long context benchmarks for base and instruct models and show that KVLinC either matches or achieves superior performance with higher KV cache compression.
    \item We develop a Triton \citep{tillet2019triton} based attention decoding kernel which along with off-the-shelf quantization kernel achieves up to $2.55\times$ faster inference and up to $3.5\times$ larger batch size with KVLinC.
\end{itemize}
\vspace{-5pt}
\section{Background}
\vspace{-5pt}
\begin{wrapfigure}{r}{0.2\textwidth} 
  \vspace{-40pt}         
  \centering
  \begin{subfigure}{0.7\linewidth}
      \centering
      \includegraphics[width=\linewidth]{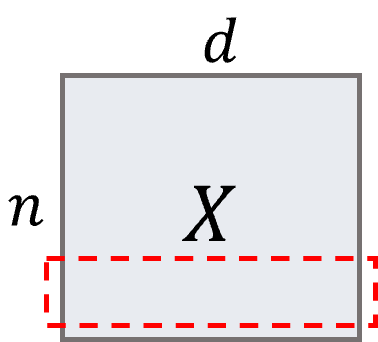}
      \caption{$Q_T(\mX)$}
  \end{subfigure}
  \centering
  \begin{subfigure}{0.7\linewidth}
      \centering
      \includegraphics[width=\linewidth]{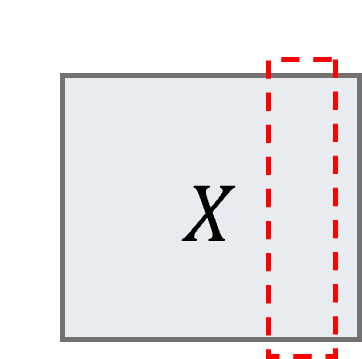}
      \caption{$Q_C(\mX)$}
  \end{subfigure}
  \vspace{-6pt}                       
  \caption{Token-wise and channel-wise quantization grouping.}
  \label{fig:quant_config}
  \vspace{-15pt}                       
\end{wrapfigure}
\textbf{Quantization.} In asymmetric integer quantization, the full-precision tensor $\mX_r$ is first mapped to an integer representation $\mX_I$ as : 
\begin{equation}
\mX_I = \Bigg\lfloor \frac{\mX_r - z}{s} \Bigg\rceil, 
\qquad 
s = \frac{\max(\mX_r) - \min(\mX_r)}{2^N - 1},
\quad 
z = \min(\mX_r),
\end{equation}  
and dequantized as : $Q(\mX) = \mX_{q} = s\mX_I + z$, where $\mX_I \in [0, 2^N - 1]$ are $N$-bit integers, $s$ is the scaling factor of quantization and $z$ is the zero-point. Quantization can be performed per tensor where $s$ and $z$ are scalars obtained for the entire tensor or, group-wise where $G$ consecutive entries share a scale factor and zero-point. Group-wise quantization reduces quantization error but requires storing multiple scale factors and zero-points. For $\mX \in \mathbb{R}^{n\times d}$, channel-wise quantization ($Q_C(\mX))$ groups entries by column $j$ and token-wise ($Q_T(\mX)$) by row $i$ as shown in Figure~\ref{fig:quant_config}. For assymetic integer quantization, the quantization error is given by \citep{peters2023qbitopt} :
\begin{equation}
    \mathbb{E}\!\left[(Q(\mX) - \mX)^2\right] = \frac{s^2}{12}
    \label{eq:quant_error}
\end{equation}

\textbf{Multi Head Attention.} A typical LLM consists of $L$ decoder layers with each layer containing a multi head attention and a feed forward network module. The multi head attention module computes attention per head in parallel with each attention head computing $\mY \in \R^{N\times d}$ from inputs $\mX \in \R^{N\times d}$ (where $N$ is sequence length and $d$ is head dimension) with query, key and value weights $\mW_q, \mW_k,\mW_v \in \R^{d\times d}$. First, we compute $\mQ=\mX\mW_q, \mK = \mX\mW_k, \mV = \mX\mW_v$, before getting attention weights $\mA$ and attention outputs $\mY$ as
\begin{equation}
    \mA_{n,i} = \frac{\exp \left( \mQ_n\mK_i^\top/\sqrt{d}\right)}{\sum_{i=1}^n \exp \left(\mQ_n\mK_i^\top/\sqrt{d} \right)}, \quad \mY_n = \sum^n_{i=1} \mA_{n,i}\mV_i, \quad \text{for $n$ in $[1,\dots,N]$}
    \label{eq:attention}
\end{equation}
The final output is obtained by concatenating $\mY$ across $h$ heads and using output projection matrix $\mW_o \in \R^{hd\times hd}$ to compute $\mO = [\mY^1, \dots \mY^h]\mW_o$. 

\textbf{LLM Inference.} LLM inference proceeds in two phases: prefill and decoding. 
In the prefill phase, per-head token embeddings have shape $\mathbb{R}^{n_p \times d}$, where $n_p$ is the prompt length. The attention computes queries, keys, and values for the prompt and caches the keys and values for subsequent steps. During decoding, the model generates $n_g$ tokens autoregressively, one at a time. At each step $t$ with $n_p < t \le n_p + n_g$, the model forms the new token embedding $\mX_t$, computes $(\mQ_t,\mK_t,\mV_t) \in \mathbb{R}^{1 \times d}$, and appends $\mK_t$ and $\mV_t$ to the cache, yielding $[\mK_0,\ldots,\mK_t]$ and $[\mV_0,\ldots,\mV_t]$. Multi-head attention then uses $\mQ_t$ to attend over the cached keys/values. With KV cache quantization, the cache stores quantized keys and values together with their scale and zero-point parameters, and these are dequantized before the attention computation.
\begin{figure}[t]
    \centering
    \includegraphics[width=\linewidth]{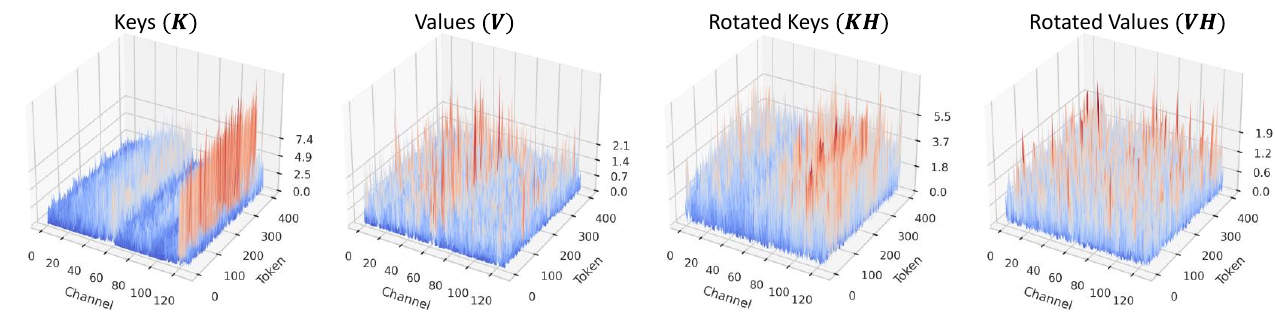}
    \caption{Distribution of key and values with and without Hadamard rotation for Qwen-2.5-3B layer 16 head 0.}
    \label{fig:kv_dist}
    \vspace{-10pt}
\end{figure}

\section{Methodology}
In this section, we introduce KVLinC, a framework for mitigating attention errors due to low-precision KV cache quantization. KVLinC integrates two complementary strategies: (i) Hadamard rotation to reduce quantization error and (ii) lightweight linear correction adapters to compensate attention distortions. We analyze axis and rotation choices for quantization, describe the design and efficiency of correction adapters, and present a custom attention kernel for accelerated decoding. These components together enable robust long-context inference at low precision with minimal overhead.

\subsection{Hadamard Rotation and KV Cache Quantization}
Key and value tensors in the KV cache follow different statistics, motivating distinct quantization schemes. As shown in Figure~\ref{fig:kv_dist}, keys contain channel-wise outliers with a few disproportionately large magnitudes, whereas values do not. KIVI~\cite{liu2024kivi} addresses this by quantizing keys channel-wise and values token-wise, yielding $\mK_q = Q_C(\mK),;\mV_q = Q_T(\mV)$. This aligns the dynamic range per column, localizing key quantization error to individual channels and matching the observed outlier structure.
\begin{figure}[t]
  \centering
  \begin{subfigure}[t]{0.3\linewidth}
    \centering
    \includegraphics[width=\linewidth]{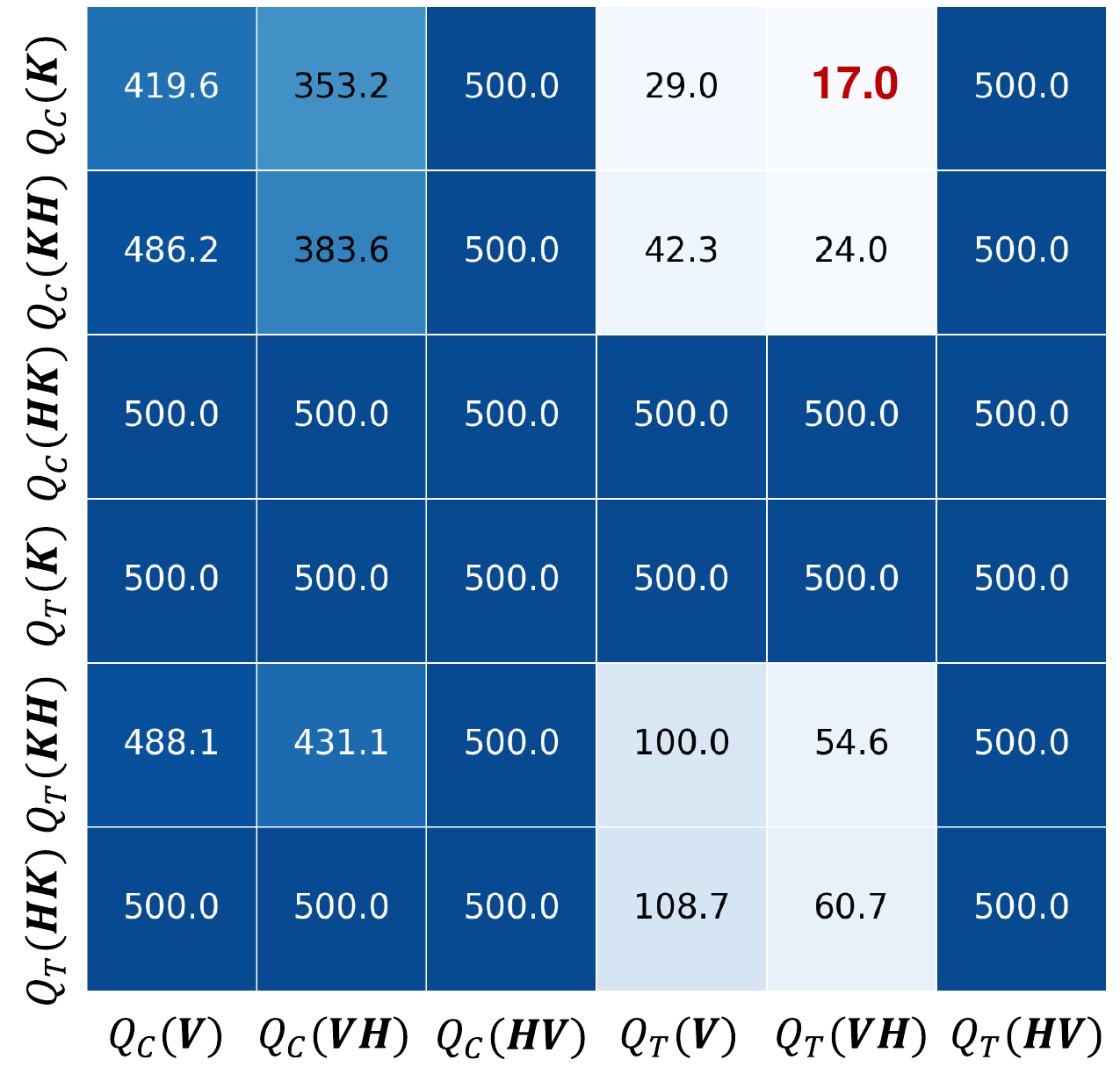}
    \caption{Llama-3.2-3B}
  \end{subfigure}
  \begin{subfigure}[t]{0.3\linewidth}
    \centering
    \includegraphics[width=\linewidth]{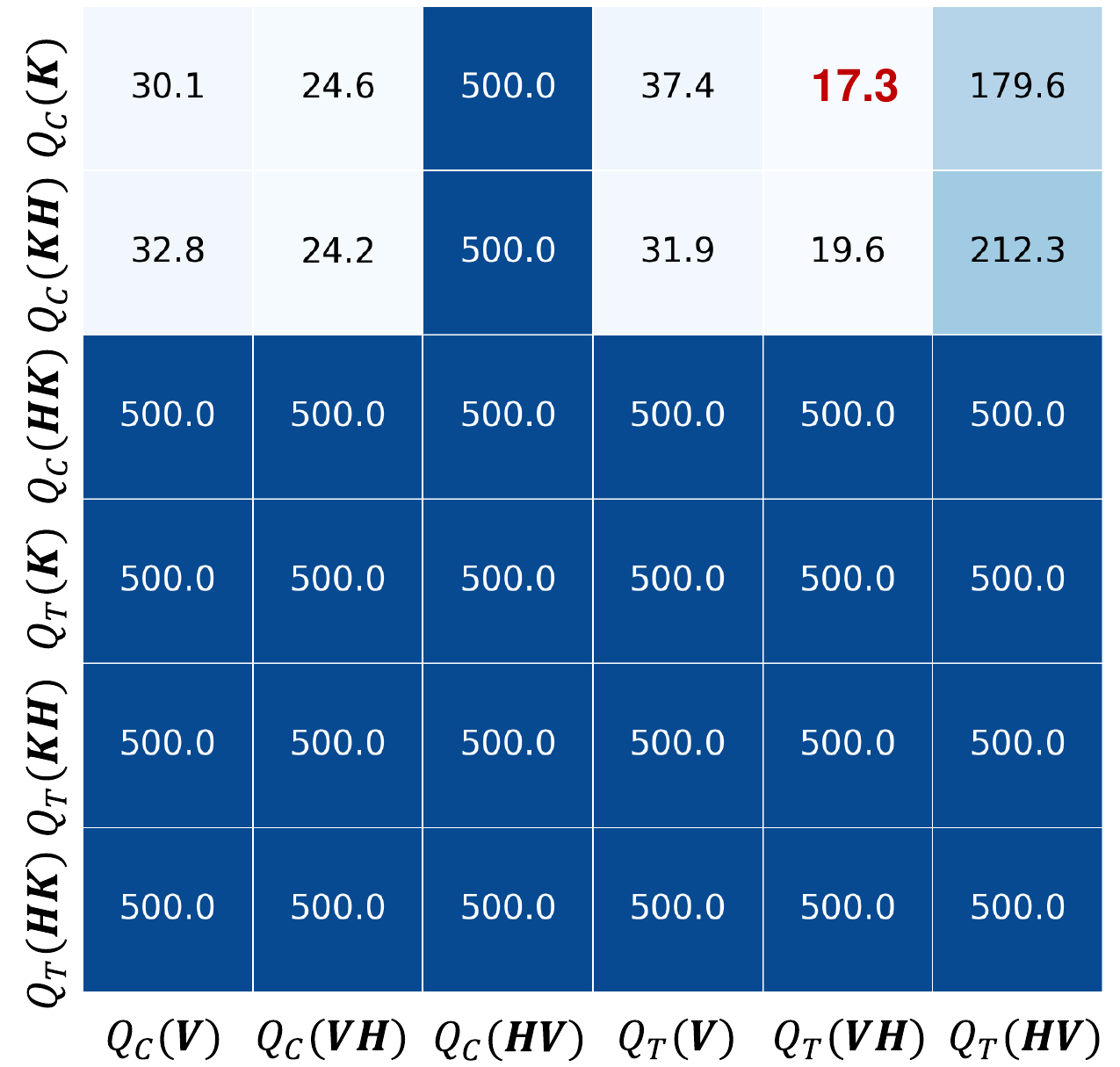}
    \caption{Qwen-2.5-3B}
  \end{subfigure}
  \begin{subfigure}[t]{0.38\linewidth}
    \centering
    \includegraphics[width=\linewidth]{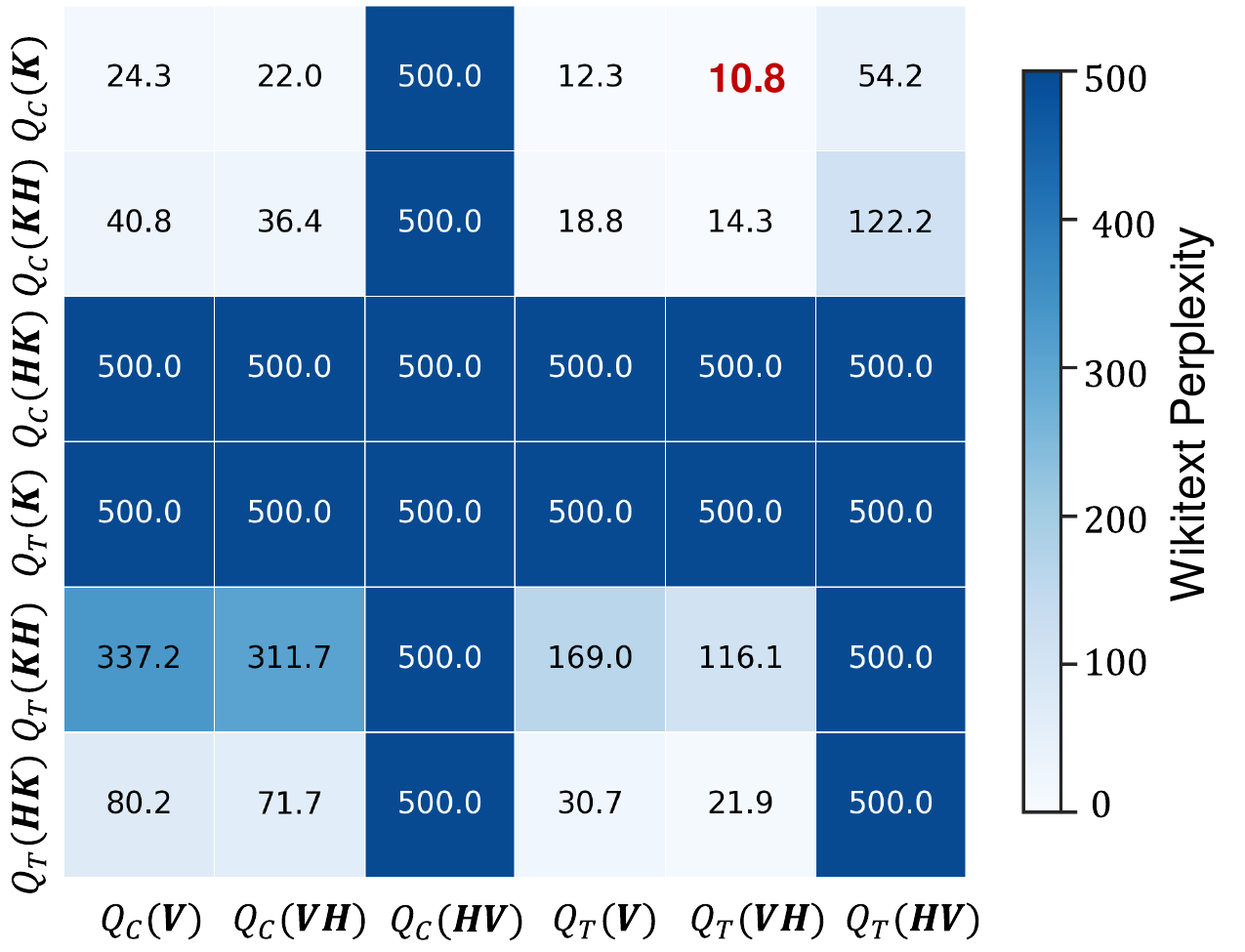}
    \caption{Qwen-3-4B}
  \end{subfigure}
  \caption{Wikitext perplexity under different 2-bit quantization configuration for key and values. Perplexity values are clipped to $500$. Quantizing raw keys channel-wise and quantizing Hadamard rotated values token-wise achieves best performance (shown in red).}
  \label{fig:quant_heatmap}
  \vspace{-15pt}
\end{figure}
In contrast, QuaRot~\citep{ashkboos2024quarot} employs a Hadamard rotation to suppress outliers and quantizes both keys and values token-wise. Denoting the Hadamard matrix by $\mH$, the quantization configuration is $\mK_q = Q_T(\mK\mH), \mV_q = Q_T(\mV\mH)$. As shown in Figure~\ref{fig:kv_dist}, the Hadamard transform equalizes key and value distributions, eliminating outliers, though its effectiveness under extreme low-precision remains untested. During dequantization, the quantized tensors $\mK_{q}$ and $\mV_{q}$ must be multiplied by $\mH^\top$, the inverse of the orthogonal Hadamard matrix, introducing additional computational overhead. While the overhead associated with values can be eliminated by merging the rotation into the projection weight matrices, keys still require online Hadamard transforms at inference time. QuaRot applies a Hadamard transform by post-multiplying keys and values before quantization ($\mK\mH$, $\mV\mH$); we also consider pre-multiplication ($\mH\mK$, $\mH\mV$). This yields a two-dimensional design space: quantization axis (channel- vs. token-wise) × Hadamard placement (pre vs. post). We ablate all combinations, quantizing $\mK$ and $\mV$ to 2-bit with group size 128, and evaluate Wikitext perplexity across three model families (Fig.~\ref{fig:quant_heatmap}). We make the following observations:

\textbf{Observation 1.} Pre-multiplying keys 
and values with a Hadamard matrix almost always
\begin{wrapfigure}{r}{0.25\textwidth} 
  \centering
  \includegraphics[width=\linewidth]{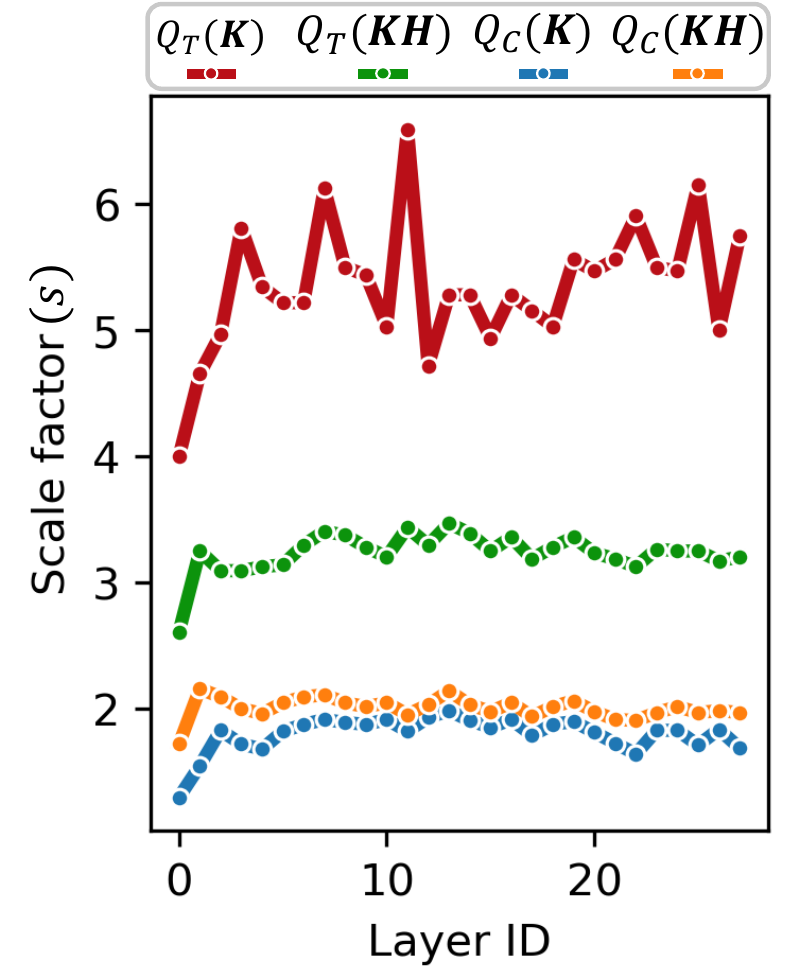}
  \vspace{-8pt}                       
  \caption{Layer-wise scaling factor for different quantization configuration of keys.}
  \label{fig:scaling_factor}
  \vspace{-20pt}                       
\end{wrapfigure}
yields worse performance compared to post-multiplication. A likely explanation is that pre-multiplication mixes tokens prior to quantization, which amplifies quantization noise and injects errors into the attention logits. In contrast, post-multiplication only rotates channels within each token, thereby preserving relative token alignment and resulting in significantly more stable performance.

\textbf{Observation 2.} At the low-precision regime under consideration, KIVI’s quantization configuration consistently outperforms QuaRot’s. QuaRot exhibits extremely high perplexity, suggesting that token-wise quantization of keys therefore still incurs large errors. To analyze quantization error (eq.~\ref{eq:quant_error}), we analyze the scaling factor for different quantization configuration of keys in Figure~\ref{fig:scaling_factor}. It shows that although, Hadamard rotation of keys reduces scaling factor and hence the quantization error with token-wise quantization, it still is much higher than channel-wise quantization of keys.

\textbf{Observation 3.} Quantizing raw keys channel-wise together with Hadamard rotated values token-wise $(\mK_q = Q_C(\mK), \mV_q = Q_T(\mV\mH))$ emerges as the optimal configuration across all model families. This even outperforms the $\mK_q = Q_C(\mK\mH), \mV_q = Q_T(\mV\mH)$ quantization scheme. This is because the application of Hadamard rotation to keys redistributes each outlier dimension, thereby increasing the scaling factor of quantization leading to higher error (Figure~\ref{fig:scaling_factor}). We therefore adopt this configuration for KVLinC quantization. Importantly, this scheme is not only optimal in terms of accuracy but also practical, as it requires no additional computational overhead for quantization or dequantization.

\subsection{Linear Correction Adapters}
To further mitigate the errors introduced in the attention operation by quantized keys, we propose correction adapters which are lightweight, trainable modules that explicitly learn to compensate for distortions in the attention distribution. Let the quantization error in keys be denoted by $\mK^e = \mK - \mK^q$. We augment the standard attention formulation with additive correction terms in both the numerator and denominator: 
\begin{equation}
    \hat{\mY_n} = 
\frac{\sum_{i=1}^n\exp \left( \mQ_n\mK^{q\top}_i/\sqrt{d}\right)\mV^q_i  +  \textcolor{blue}{\sum_{i=1}^nf(\mQ_n, \mK^e_i)\mV^q_i}}{\sum_{i=1}^n \exp \left(\mQ_n\mK^{q\top}_i/\sqrt{d} \right) + \textcolor{blue}{\sum_{i=1}^nf(\mQ_n, \mK^e_i)}}.
\end{equation}
Given a query, these correction terms add residual attention weights corresponding to the error induced by quantization. 
By reparameterizing the correction term additively, we obtain a lightweight approximation that captures the dominant error while remaining computationally efficient. Let the correction adapters $\phi_q, \phi_k : \mathbb{R}^{d} \mapsto \mathbb{R}^{D}$ be the trainable feature maps. We define the correction term as the dot product of query and key error feature maps: $f(\mQ_n, \mK^e_i) = \phi_q(\mQ_n)\phi_k(\mK^e_i)^\top$. This allows the numerator of the correction term to be written as $\phi(\mQ_n)\sum_{i=1}^n\phi_k(\mK^e_i)^\top \mV^q_i$, and the denominator as $\phi_q(\mQ_n)\sum_{i=1}^n\phi_k(\mK^e_i)^\top$. \begin{wrapfigure}{r}{0.32\textwidth} 
  \vspace{-10pt}                        
  \centering
  \includegraphics[width=0.30\textwidth]{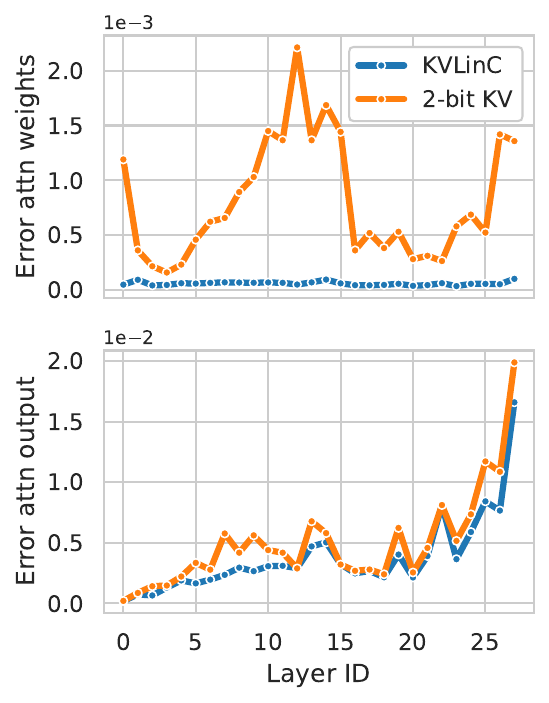}
  \vspace{-8pt}                       
  \caption{Layer-wise attention error from KV-cache quantization: (top) MSE between full-precision and quantized attention weights; (bottom) MSE between outputs. KVLinC (blue) consistently reduces error versus 2-bit KV (orange).}
  \label{fig:attn_error}
  \vspace{-29pt}                       
\end{wrapfigure}
With $\mS_0 = \mathbf{0}$ and $\mP_0 = \mathbf{0}$, we compute attention as, 
\begin{equation}
\label{eq:kvlinc_attention}
\hat{\mY_n} = 
\frac{\sum_{i=1}^n\exp \left( \mQ_n(\mK^q_i)^\top/\sqrt{d}\right)\mV^q_i  + \textcolor{blue}{\phi_q(\mQ_n)\mS_n}}{\sum_{i=1}^n \exp \left( \mQ_n(\mK^q_i)^\top/\sqrt{d}\right) + \textcolor{blue}{\phi_q(\mQ_n)\mP_n}}, 
\end{equation}
for $\textcolor{blue}{\mS_n = \mS_{n-1} + \phi_k(\mK^e_n)^\top \mV_n}$ and $\textcolor{blue}{\mP_n = \mP_{n-1} + \phi_n(\mK^e_n)}$. 
This recurrent formulation transforms the quadratic accumulation of correction terms into linear-time updates, allowing error compensation to scale efficiently with sequence length. The cost of error correction is $\mathcal{O}(ndD)$ in time and memory during prefill, and only $\mathcal{O}(dD)$ per step during decoding. At decoding time, the cache stores the quantized keys and values along with the correction states $\mS_n \in \R^{d\times D}$ and $\mP_n \in \R^{D}$. The additional memory cost is constant with respect to sequence length, making the correction adapters highly efficient.  Following LolCats \cite{zhang2024lolcats}, we choose the feature maps $\phi$ as
\begin{equation}
    \phi(\mX) = 
\left[
    \operatorname{softmax}(\mX\mW_1), \operatorname{softmax}(\mX\mW_2)
\right] \in \mathbb{R}^D
\end{equation}
with learnable weights $\mW_1, \mW_2 \in \mathbb{R}^{d\times D/2}$.  The trainable feature maps add less than $1\%$ parameter overhead. The weights are trained such that the full-precision attention weights $\mA_{n,i}$ (eq. \ref{eq:attention}) match the corrected quantized attention weights $\hat{\mA}_{n,i}$ :
\[
    \hat{\mA}_{n,i} = 
\frac{\exp \left( \mQ_n\mK^{q\top}_i/\sqrt{d}\right)  + \phi_q(\mQ_n)\phi_k(\mK^e_i)^\top}{\sum_{i=1}^n \exp \left( \mQ_n\mK^{q\top}_i/\sqrt{d}\right) + \phi_q(\mQ_n)\phi_k(\mK^e_i)^\top}
\]
Using a calibration dataset, we optimize the feature map parameters to reduce the cross-entropy loss between $\mA_{n,i}$ and $\hat{\mA}_{n,i}$. As shown in Figure~\ref{fig:attn_error}, after training, the error between quantized attention and full precision attention is minimized. Thus, correction adapters enable quantized attention to closely match full-precision distributions.

\subsection{System Level Implementation}
Improving end-to-end performance with KV-cache quantization requires custom kernels to (1) quantize the cache and (2) run attention directly on quantized operands. We adopt the quantization kernel from KIVI \citep{liu2024kivi}, which quantizes the KV cache to 2 bits and bit-packs 16 elements into a single 32-bit word. To accelerate decoding, we implement a custom attention kernel in
Triton \citep{tillet2019triton}. In the spirit of FlashAttention \citep{dao2023flashattention}, the kernel streams blocks of keys and values from off-chip High Bandwidth Memory (HBM) to on-chip Static Random Access Memory (SRAM), performs dequantization and the attention computations on chip, and writes partial outputs. Because the decoding phase exposes limited parallelism, we parallelize across KV blocks: each block produces a partial sum of the attention output in parallel, the
partial sums are reduced to form the final attention output, and we then apply the KVLinC linear correction. Before running attention, we compute the linear-correction states
$\mC_n$ and $\mC_d$ for the numerator and denominator, respectively: $\mC_n = \phi_q(\mQ_n)\mS_n,\mC_d = \phi_q(\mQ_n)\mP_n.$
These states, together with the (de)quantized attention operands, are passed to the decoding algorithm in Algorithm~\ref{alg:kvlincdecoding}. Following KIVI \cite{liu2024kivi}, we quantize the KV cache only after the attention computation; consequently, the prefill phase remains floating-point and can be accelerated with FlashAttention itself.

\begin{algorithm}[t]
\caption{KVLinC forward pass (single decode step)}
\label{alg:kvlincdecoding}
\begin{algorithmic}[1]
\Require $\mQ \in \mathbb{R}^{1\times d}, \mK_I\in\mathbb{R}^{N/16\times d},\mZ_k,\mS_k\in\mathbb{R}^{N/G\times d}, \mV_I\in\mathbb{R}^{N\times d/16}, \mZ_v,\mS_v\in\mathbb{R}^{N\times d/G}$, $\mC_n, \mC_d \in \R^{1\times d}, G$ (quantization group size).
\Ensure Output $\mY \in \mathbb{R}^{1\times d}$.
\State Divide $\mK_I,\mZ_k,\mS_k,\mV_I,\mZ_v,\mS_v$ in $T=\left\lceil N/G \right\rceil$ blocks: $\mK^1_{I},\dots,\mK^T_{I}$ of size $\frac{G}{16}\times d$ each, $\mS^1_{k},\mZ^1_{k}\dots,\mZ^T_{k},\mS^T_{k}$ of size $1\times d$ each, $\mV^1_{I}\dots \mV^T_{I}$ of size $G\times \frac{d}{16}$ each and $\mS^1_{v},\mZ^1_{v}\dots \mS^T_{v},\mZ^T_{v}$ of size $G\times \frac{d}{G}$ each.

\State Create empty softmax state $\mY_s \in \R^{T\times d}, \vl, \vm \in \R^T$ 

\State Load $\mQ$ from HBM to SRAM.

\State \textbf{parallel For} $j=1$ to $T$ \Comment{Parallelized across KV blocks}
  \State \quad Load $\mK^j_I, \mS^j_k, \mZ^j_k, \mV^j_I, \mS^j_v, \mZ^j_v$ from HBM to SRAM.
  \State \quad On chip, dequantize keys: $\mK^{j\top}_{q} = \operatorname{unpack}(\mK_I^{j\top}) \odot \mS_k^{j} + \mZ_k^{j}$.
  \State \quad On chip, compute $\mS^j=s\cdot \mQ\mK^\top_{q}$.
  \State \quad On chip, compute $m^j = \operatorname{rowmax}(\mS^j), \quad \mE^j = \exp(\mS^j-m), \quad l^j = \operatorname{rowsum}(\mE^j). $
  \State \quad On chip, dequantize values : $\mV^j_{q} = \operatorname{unpack}(\mV^j_I) \odot \mS_v^j + \mZ_v^j$.
  \State \quad On chip, compute : $\mY^j_s=\mE^j\mV^j_{q} \in \R^{1\times d}$.
  \State \quad Write $\mY^j_s$, $m^j$ and $l^j$ to HBM.
\State \textbf{end parallel For}
\State $\vw = \exp(\vm - \max(m))$
\State $\vn = \operatorname{rowsum}(\mY_s\cdot \vw) + \mC_n$, \quad $\vd = (\vl\cdot\vw) + \mC_d$ \Comment{attention numerator and denominator}
\State $\mY = \vn/\vd$ 
\end{algorithmic}
\end{algorithm}

\section{Experiments}
\vspace{-10pt}
In this section, we benchmark KVLinC against competitive baselines. First we provide results of the algorithm and then we provide end to end hardware efficiency improvements provided by KVLinC.
\vspace{-8pt}
\subsection{Setup}
\label{sec:setup}
\textbf{Models, tasks, datasets and baselines.} We evaluate KVLinC on the Llama-3 \citep{llama3.2meta,llama3meta,touvron2023llama2}, Qwen-2.5 \citep{yang2024qwen2}, and Qwen-3 \citep{yang2025qwen3technicalreport} model families, chosen to test robustness of linear correction adapters under different architectural settings (Qwen-2.5 uses bias in query/key projections; Qwen-3 applies layernorm after them). We compare against KIVI \citep{liu2024kivi}, QuaRot \citep{ashkboos2024quarot}, ResQ \citep{saxena2024resq}, and Gear \citep{kang2024gearefficientkvcache}. Since ResQ and Gear retain portions of the KV cache in high precision, we align their design point with KVLinC’s average precision: ResQ keeps $3.125\%$ of channels in 16-bit, and Gear uses rank-2 quantization error. All methods quantize the KV cache to 2-bits with group size 128, while storing the most recent 128 tokens in full precision. We evaluate both base and instruction-tuned models. For base models, we measure perplexity on Wikitext \citep{merity2016wikitext} (2k sequence length, autoregressive generation with compressed KV cache), exact match accuracy on 5-shot GSM8K \citep{cobbe2021gsm8k}, and average accuracy on Big-Bench Hard (BBH) \citep{suzgun2022bigbenchhard}. For instruction-tuned models, we report results on long-context benchmarks: RULER \citep{hsieh2024ruler}, LongBench \citep{bai2023longbench}, and IF-Eval \citep{zhou2023ifeval}. LongBench follows the setup in KIVI, while other benchmarks use the lm-evaluation-harness \citep{lm-eval-harness}.

\textbf{Implementation Details} We implement KVLinC in PyTorch \citep{pytorch-Neurips2019} using HuggingFace Transformers \citep{huggingface-arxiv2019}. We set rank of correction adapters as $D=256$, adding $<1\%$ extra parameters to the LLM. For base models, adapters are trained on Alpaca dataset \citep{taori2023alpaca} using Adam \citep{kingma2017adammethodstochasticoptimization} optimizer with learning rate 0.01, sequence length 3k, batch size 24, for 500 steps. For instruction-tuned models, training uses RedPajama dataset \citep{weber2024redpajama}, sequence length 8k, batch size 8, for 1500 steps with Adam optimizer. Training Llama-3.1-8B on Alpaca takes ~2 hours, and Llama-3.2-3B on RedPajama takes ~11 hours on 4×NVIDIA H200 GPUs.
\begin{table}[t!]
\centering
\caption{Results of base LLMs on Wikitext perplexity (2k sequence length), 5-shot GSM8K and BBH. Average KV cache precision is computed considering the scaling factors and zero points along with components used to compensate for quantization error. $\uparrow$ higher is better, $\downarrow$: lower is better. \textsuperscript{*}Upper bound performance.}
\label{tab:results_base}
\resizebox{\textwidth}{!}{
\begin{tabular}{c|c|ccc|ccc|ccc}
\hline
 & KV & \multicolumn{3}{c|}{Llama-2-7B} & \multicolumn{3}{c|}{Llama-3.2-3B} & \multicolumn{3}{c}{Llama-3.1-8B} \\ \cline{3-11} 
\multirow{-2}{*}{Method} & Cache & Wikitext$\downarrow$&GSM8K$\uparrow$&BBH$\uparrow$& Wikitext$\downarrow$&GSM8K$\uparrow$&BBH$\uparrow$& Wikitext$\downarrow$&GSM8K$\uparrow$&BBH$\uparrow$ \\ \hline
FP16\textsuperscript{*} & 16 & 5.5 & 14.3 & 39.9 & 7.8 & 25.6 & 47.0 & 6.2 & 49.7 & 62.7 \\ 
KIVI & 2.46 & 5.9 & 10.6 & 30.5 & 11.0 & 11.8 & 25.0 & 7.8 & 34.1 & 44.2 \\
Quarot & 2.46 & 5.8 & 9.8 & 29.4 & 9.7 & 9.9 & 21.5 & 7.3 & 26.8 & 34.3 \\
ResQ & 2.91 & \textbf{5.7} & 10.8 & \textbf{32.6} & \textbf{8.7} & 14.1 & 31.5 & \textbf{6.8} & 36.2 & 42.7 \\
Gear-L & 2.96 & 5.8 & 10.8 & 30.0 & 10.0 & \textbf{16.4} & 28.3 & 7.3 & 38.8 & 46.7 \\
\rowcolor[HTML]{CEEFF8} 
KVLinC & 2.71 & \textbf{5.7} & \textbf{11.0} & 31.1 & 9.4 & \textbf{16.4} & \textbf{32.7} & 7.1 & \textbf{40.9} & \textbf{48.6} \\ \hline
 & KV & \multicolumn{3}{c|}{Qwen2.5-1.5B} & \multicolumn{3}{c|}{Qwen2.5-3B} & \multicolumn{3}{c}{Qwen2.5-7B} \\ \cline{3-11} 
\multirow{-2}{*}{Method} & Cache & Wikitext$\downarrow$&GSM8K$\uparrow$&BBH$\uparrow$& Wikitext$\downarrow$&GSM8K$\uparrow$&BBH$\uparrow$& Wikitext$\downarrow$&GSM8K$\uparrow$&BBH$\uparrow$ \\ \hline
FP16\textsuperscript{*} & 16 & 9.3 & 61.5 & 43.9 & 8.0 & 69.4 & 55.1 & 6.8 & 81.1 & 69.4 \\ 
KIVI & 2.46 & 16.5 & 26.9 & 17.9 & 9.7 & 46.1 & 32.7 & 11.2 & 71.0 & 45.3 \\
Quarot & 2.46 & 7268.2 & 0.1 & 0.0 & 783.0 & 0.0 & 0.0 & 3380.0 & 0.1 & 0.0 \\
ResQ & 2.91 & 13.2 & 10.6 & 22.1 & 9.1 & 47.2 & 39.2 & 10.6 & 35.6 & 47.9 \\
Gear-L & 2.96 & 14.0 & 32.2 & 21.7 & 9.3 & 47.4 & 34.1 & 10.6 & \textbf{71.8} & 49.5 \\
\rowcolor[HTML]{CEEFF8} 
KVLinC & 2.71 & \textbf{13.0} & \textbf{36.3} & \textbf{23.6} & \textbf{8.9} & \textbf{47.6} & \textbf{35.3} & \textbf{10.5} & 71.2 & \textbf{50.1} \\ \hline
 & KV & \multicolumn{3}{c|}{Qwen3-1.7B-Base} & \multicolumn{3}{c|}{Qwen3-4B-Base} & \multicolumn{3}{c}{Qwen3-8B-Base} \\ \cline{3-11} 
\multirow{-2}{*}{Method} & Cache & Wikitext$\downarrow$&GSM8K$\uparrow$&BBH$\uparrow$& Wikitext$\downarrow$&GSM8K$\uparrow$&BBH$\uparrow$& Wikitext$\downarrow$&GSM8K$\uparrow$&BBH$\uparrow$ \\ \hline
FP16\textsuperscript{*} & 16 & 9.4 & 69.3 & 53.2 & 7.9 & 76.0 & 71.3 & 7.0 & 82.3 & 77.3 \\ 
KIVI & 2.46 & 11.2 & 48.4 & 30.5 & 9.1 & 67.5 & 49.9 & 7.7 & 78.6 & 58.6 \\
Quarot & 2.46 & 1963.3 & 0.0 & 0.0 & 755.3 & 0.1 & 0.0 & 202.3 & 17.5 & 20.8 \\
ResQ & 2.9 & 12.2 & 20.4 & 18.8 & 9.0 & 48.9 & 51.0 & 7.8 & 71.7 & 58.5 \\
Gear-L & 2.96 & 10.7 & 47.5 & 33.2 & 8.8 & 66.9 & 55.1 & 7.6 & 78.6 & \textbf{63.2} \\
\rowcolor[HTML]{CEEFF8} 
KVLinC & 2.71 & \textbf{10.4} & \textbf{53.9} & \textbf{35.5} & \textbf{8.6} & \textbf{67.6} & \textbf{55.2} & \textbf{7.5} & \textbf{78.9} & 61.7 \\ \hline
\end{tabular}
}
\end{table}
\begin{table}[t!]
\centering
\caption{Results of Instruct LLMs on long context and instruction following tasks. Taskwise accuracy can be found in Appendix~\ref{sec:app_longbench},\ref{sec:app_ruler}.\textsuperscript{*}Upper bound performance.}
\label{tab:results_instruct}
\resizebox{0.9\textwidth}{!}{
\begin{tabular}{c|c|c|cc|c|cc}
\hline
 &  &  & \multicolumn{2}{c|}{RULER} &  & \multicolumn{2}{c}{IF-Eval} \\ \cline{4-5} \cline{7-8} 
\multirow{-2}{*}{Model} & \multirow{-2}{*}{Method} & \multirow{-2}{*}{KV Cache} & 4k & 8k & \multirow{-2}{*}{LongBench} & inst-strict & prompt-strict \\ \hline
 & FP16\textsuperscript{*} & 16 & 92.5 & 88.1 & 40.4 & 79.3 & 71.2 \\ 
 & KIVI & 2.46 & 76.7 & 70.3 & \textbf{39.4} & 74.6 & 64.9 \\
\multirow{-4}{*}{Llama-3.2-3B-Instruct} & \cellcolor[HTML]{CEEFF8}KVLinC & \cellcolor[HTML]{CEEFF8}2.71 & \cellcolor[HTML]{CEEFF8}\textbf{80.8} & \cellcolor[HTML]{CEEFF8}\textbf{73.6} & \cellcolor[HTML]{CEEFF8}\textbf{39.4} & \cellcolor[HTML]{CEEFF8}\textbf{76.3} & \cellcolor[HTML]{CEEFF8}\textbf{67.5} \\ \hline
 & FP16\textsuperscript{*} & 16 & 90.3 & 85.0 & 31.4 & 68.9 & 58.8 \\ 
 & KIVI & 2.46 & 49.5 & 41.0 & 28.0 & 62.7 & 52.5 \\
\multirow{-4}{*}{Qwen-2.5-3B-Instruct} & \cellcolor[HTML]{CEEFF8}KVLinC & \cellcolor[HTML]{CEEFF8}2.71 & \cellcolor[HTML]{CEEFF8}\textbf{60.9} & \cellcolor[HTML]{CEEFF8}\textbf{51.1} & \cellcolor[HTML]{CEEFF8}\textbf{28.2} & \cellcolor[HTML]{CEEFF8}\textbf{66.0} & \cellcolor[HTML]{CEEFF8}\textbf{56.8} \\ \hline
 & FP16\textsuperscript{*} & 16 & 92.7 & 88.6 & 31.9 & 47.6 & 33.6 \\ 
 & KIVI & 2.46 & 83.7 & 79.9 & \textbf{31.2} & 44.8 & 31.8 \\
\multirow{-4}{*}{Qwen-3-4B-Instruct} & \cellcolor[HTML]{CEEFF8}KVLinC & \cellcolor[HTML]{CEEFF8}2.71 & \cellcolor[HTML]{CEEFF8}\textbf{86.2} & \cellcolor[HTML]{CEEFF8}\textbf{82.4} & \cellcolor[HTML]{CEEFF8}31.0 & \cellcolor[HTML]{CEEFF8}\textbf{45.7} & \cellcolor[HTML]{CEEFF8}\textbf{32.5} \\ \hline
\end{tabular}
}
\end{table}
\subsection{Main Results}
\begin{figure}[t]
  \centering
  \begin{minipage}[t!]{0.46\textwidth}
    \centering
    \captionof{table}{Performance with applying Hadamard rotation and linear correction in isolation on Llama family. $\uparrow$ higher is better, $\downarrow$: lower is better.}
    \label{tab:diff_components}
    \resizebox{\linewidth}{!}{
    \begin{tabular}{cc|cc}
    \hline
    Model & Method & Wikitext$\downarrow$ & GSM8K$\uparrow$ \\ \hline
    \multirow{4}{*}{3.1-8B} & KIVI & 7.8 & 34.1 \\
     & KIVI + LinC & 7.3 & 38.4 \\
     & $Q_C(\mK), Q_T(\mV\mH)$ & 7.2 & 36.9 \\
     & KVLinC & 7.1 & 40.9 \\ \hline
    \multirow{4}{*}{3.2-3B} & KIVI & 11.0 & 11.8 \\
     & KIVI + LinC & 9.8 & 14.5 \\
     & $Q_C(\mK), Q_T(\mV\mH)$ & 9.7 & 13.9 \\
     & KVLinC & 9.4 & 16.4 \\ \hline
    \end{tabular}
    }
  \end{minipage}%
  \hfill%
  \begin{minipage}[!t]{0.48\textwidth}
    \centering
    \captionof{table}{Impact on wikitext PPL with applying KVLinC to different decoder layer blocks. Applying KVLinC to earlier decoder layers provides greater improvements.}
    \label{tab:block-impr-table}
    \resizebox{\linewidth}{!}{
    \begin{tabular}{c|cc}
    \hline
    KVLinC & \multicolumn{2}{c}{Improvement over KIVI (\%)} \\ \cline{2-3} 
     Layers & \multicolumn{1}{c|}{Qwen-2.5-1.5B} & Qwen-3-1.7B-Base \\ \hline
    \texttt{[0-9]} & \multicolumn{1}{c|}{7.96} & 3.03 \\
    \texttt{[9-18]} & \multicolumn{1}{c|}{4.35} & 2.44 \\
    \texttt{[18-27]} & \multicolumn{1}{c|}{2.73} & 1.20 \\ \hline
    \texttt{[0-13]} & \multicolumn{1}{c|}{10.55} & 3.75 \\
    \texttt{[14-27]} & \multicolumn{1}{c|}{4.29} & 2.33 \\
    \texttt{[0-27]} & \multicolumn{1}{c|}{16.82} & 5.27 \\ \hline
    \end{tabular}
    }
  \end{minipage}%
\end{figure}

\textbf{Results on Base Models.} We evaluate the base LLMs of various sizes of Llama, Qwen-2.5, and Qwen-3 model families on perplexity (PPL) on Wikitext at 2k sequence length, 5-shot GSM8K, and BBH benchmark. The results are presented in Table~\ref{tab:results_base}. KVLinC manages to outperform or match the performance of strong baselines at lower KV cache bitwidth. Compared to Gear, KVLinC achieves upto $6.4\%$ improvements on GSM8K and upto $2.3\%$ improvements on BBH benchmark. Greater improvements are observed for smaller-sized models. For the Qwen-2.5 and Qwen-3 family of models, QuaRot fails to produce meaningful results, showcasing that per token quantization strategy for both keys and values is sub-optimal. ResQ adopts the same quantization configuration as QuaRot but keeps important channels in high precision, enabling improved results. Since calibration for ResQ is done on Wikitext itself, it achieves surprisingly low Wikitext PPL on Llama models. KVLinC instead involves calibration on out-of-domain Alpaca dataset and does not overfit to any evaluation benchmarks. 

\textbf{Results on Instruct models.} We evaluate the instruction tuned LLMs of Llama-3.1, Qwen-2.5 and Qwen-3 model families on RULER (4k and 8k sequence length), LongBench and IF-eval benchmarks. The results are presented in Table~\ref{tab:results_instruct}. KVLinC outperforms KIVI on all the presented models on RULER and IF-eval tasks. For the Qwen-2.5-3B instruct model, KVLinC achieves more than $10\%$ improvement on RULER tasks and upto $4.3\%$ on IF-eval tasks. For LongBench, quantization of KV cache impacts final accuracy by a small amount and the performance of both KIVI and KVLinC is comparable. 
\vspace{-5pt}
\subsection{Analysis}
\vspace{-5pt}
\begin{wrapfigure}{r}{0.28\textwidth} 
  \vspace{-25pt}                         
  \centering
  \includegraphics[width=\linewidth]{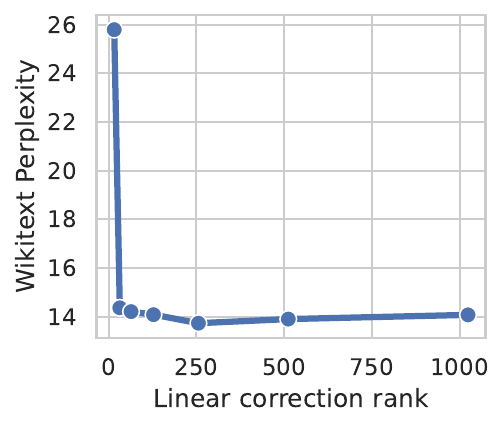}
  \caption{Linear correction rank ($D$) vs. perplexity.}
  \label{fig:rank_linc}
  \vspace{-10pt}                        
\end{wrapfigure}
\textbf{Impact of different components.} Further, we analyse how the complementary strategies presented in KVLinC perform in isolation. To achieve this, we apply the linear correction states to KIVI (KIVI+LinC) and compare with a baseline which does channel-wise quantization on raw keys and token-wise quantization on hadamard rotated values. The results are presented in Table~\ref{tab:diff_components}. For both Llama-3.1-8B and Llama-3.2-3B, applying linear correction provides improvements in Wikitext perplexity and 5-shot GSM8K accuracy. Similarly, opting for Hadamard based quantization for values improves performance over KIVI. While combining the two complementary techniques enables KVLinC to achieve further gains in performance. 

\textbf{Layerwise insights.} To better understand where KVLinC provides the most benefit, we selectively apply it to different subsets of decoder layers while using KIVI’s quantization strategy for the remaining layers. On Qwen-2.5-1.5B and Qwen-3-1.7B-Base (both with 28 decoder layers), we observe that applying KVLinC to earlier layers yields greater improvements than applying it to the same number of later layers. As shown in Table~\ref{tab:block-impr-table}, the Wikitext perplexity improvements (relative to KIVI) are consistently higher when KVLinC is applied to the initial layers. For example, applying KVLinC to the first 10 decoder layers achieves an average $3.5\%$ improvement over applying it to the last 10 layers. This finding highlights a key insight: the initial decoder layers play a more critical role under KV cache quantization.

\textbf{Dimension of Linear correction states.} The rank of the linear correction states $D$ controls the representational capacity of the feature maps, but higher ranks also increase overhead. As shown in Figure~\ref{fig:rank_linc}, Wikitext perplexity improves with larger ranks up to $D=256$, beyond which gains saturate. We therefore select $D=256$ as the optimal balance between accuracy and efficiency.
\vspace{-5pt}
\subsection{Hardware Speedup}
\begin{figure}[t]
  \centering
  \begin{subfigure}[t]{0.49\linewidth}
    \centering
    \includegraphics[width=\linewidth]{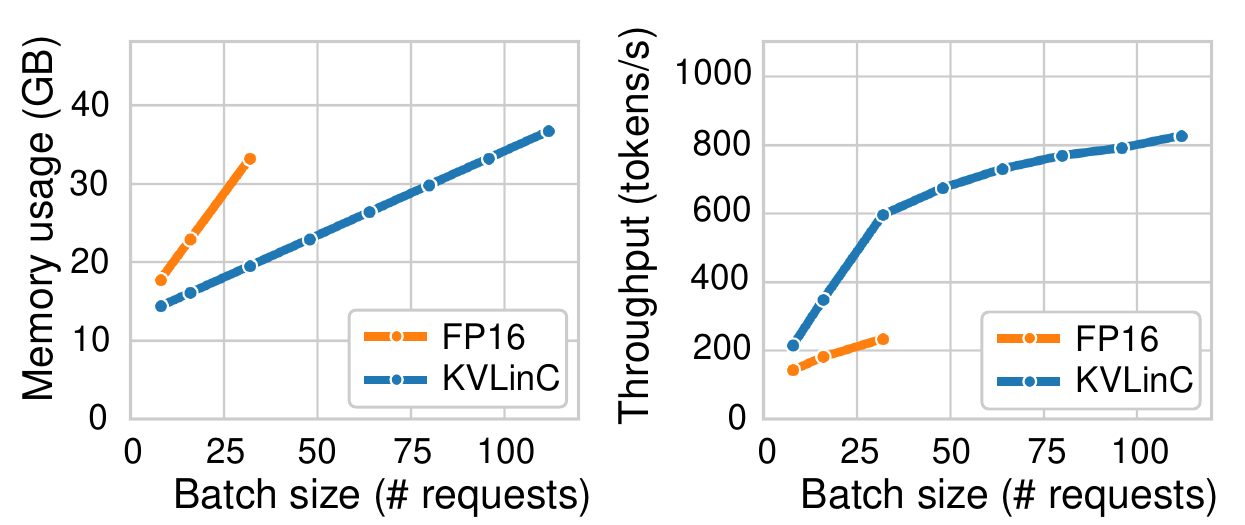}
    \caption{Llama-2-7B}
    \label{fig:llama-2-7b-perf}
  \end{subfigure}
  \begin{subfigure}[t]{0.49\linewidth}
    \centering
    \includegraphics[width=\linewidth]{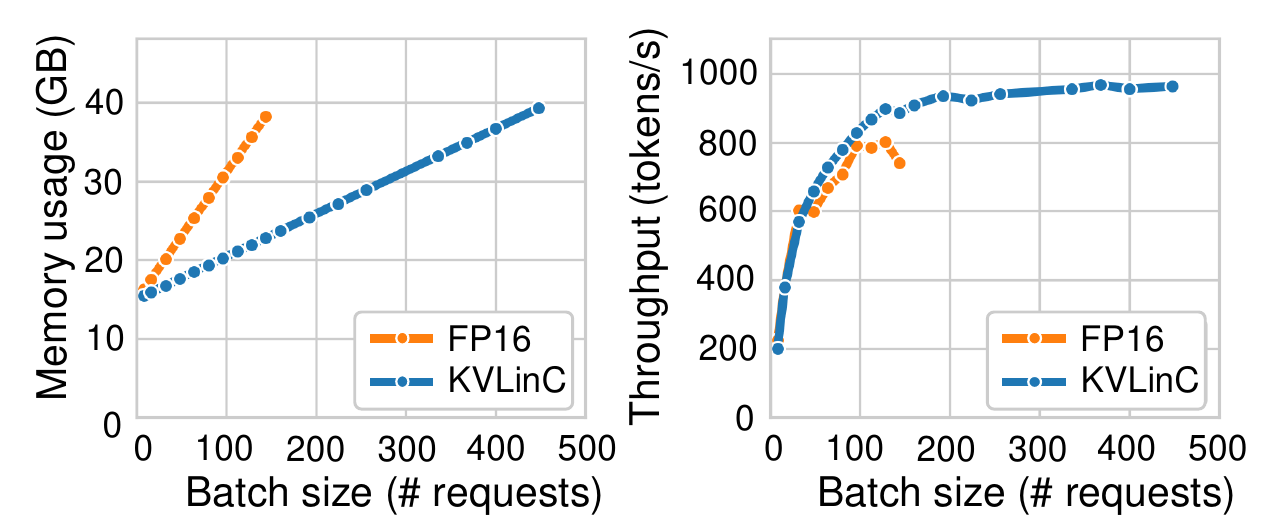}
    \caption{Llama-3.1-8B}
    \label{fig:llama-3-8b-perf}
  \end{subfigure}
  \caption{End to end memory usage and throughput (tokens/s) on NVIDIA-A40 with varying batch sizes at prompt length of $256$ and $1024$ generated tokens for (a) Llama-2-7B and (b) Llama-3.1-8B.}
  \label{fig:hardware_perf}
\end{figure}

\vspace{-5pt}
We evaluate the end-to-end speedup of KVLinC to highlight the combined impact of KV cache quantization and our custom compute kernel. Specifically, we benchmark Llama-2-7B and Llama-3.1-8B using a prompt length of $256$ tokens and generating $1024$ output tokens, progressively increasing the batch size. Experiments are conducted on a single NVIDIA A40 (48 GB) GPU, measuring both memory usage and throughput (tokens per second). We compare KVLinC against FlashAttention-2 \cite{dao2023flashattention} with a 16-bit floating-point KV cache. As shown in Figure~\ref{fig:hardware_perf}, quantizing the KV cache enables significantly larger batch sizes without exhausting memory. In particular, KVLinC supports up to $3.1\times$ more requests on Llama-3.1-8B and $3.5\times$ more requests on Llama-2-7B. Moreover, for Llama-2-7B, KVLinC delivers up to $2.55\times$ faster inference at batch size 32, beyond which FlashAttention becomes infeasible due to out-of-memory errors. For Llama-3.1-8B, the gains are more modest, with KVLinC achieving $1.2\times$ speedup at batch size $144$. This discrepancy arises from architectural differences: unlike Llama-3.1-8B, Llama-2-7B does not employ grouped query attention (GQA), resulting in a substantially larger KV cache that amplifies the benefits of our method.

\vspace{-5pt}
\section{Related Works}
\vspace{-5pt}
\textbf{KV Cache Quantization} The dynamic nature of KV caching introduces unique challenges for quantization, where both quantization and dequantization speed is critical. A variety of strategies have been explored across different granularities. ZipCache \citep{he2024zipcache} and WKVQuant \citep{yue2024wkvquant} adopt channel-separable, token-wise quantization, while KIVI \citep{liu2024kivi} applies channel-wise quantization to keys and token-wise quantization to values. In contrast, KVQuant \citep{hooper2024kvquant} and PolarQuant \citep{han2025polarquant} use non-linear quantization schemes to reduce error. QJL \citep{zandieh2025qjl} introduces a specialized Johnson–Lindenstrauss transform for key tensors combined with per-token quantization of values. Other methods combine quantization with decomposition: Palu \citep{chang2024palu} and EigenAttention \citep{saxena2024eigen} integrate low-rank factorization with quantization. Several approaches further mitigate quantization error by leveraging advanced transformations or error modeling. QuaRot \citep{ashkboos2024quarot} and SpinQuant \citep{liu2024spinquant} use Hadamard transforms to improve quantization robustness. ResQ \citep{saxena2024resq} preserves salient channels in higher precision, whereas GEAR \citep{kang2024gearefficientkvcache} maintains a low-rank approximation of the quantization error. Finally, MiKV \citep{yang2024mikv}, QAQ \citep{dong2024qaq}, and SKVQ \citep{duanmu2024skvq} explore variable bit-width schemes to balance accuracy with memory savings.

\textbf{Linear Attention} A large body of prior work has explored more efficient sequence modeling modules as alternatives to softmax attention in transformers, often by pretraining architectures from scratch. Within this line, numerous linear attention approaches have been proposed \cite{choromanski2020performer, katharopoulos2020transformersarernns, xiong2021nystromformer, yang2023gatedlinearattention}. More recently, several efforts focus on post-training conversion of softmax-attention transformers into linear counterparts. For example, Lolcats \citep{zhang2024lolcats} employs advanced linear feature map design combined with attention distillation, while Liger \citep{lan2025liger} incorporates gated recurrence to achieve this transition. Pushing further, LoLA \citep{mcdermott2025lolalowranklinearattention} and Based \citep{arora2025based} adopt hybrid strategies that combine linear attention with selective application of exact softmax attention on subsets of keys and values, thereby improving accuracy while retaining efficiency.

\vspace{-5pt}
\section{Conclusion}
\vspace{-5pt}
In this work, we introduced \emph{KVLinC}, a framework designed to mitigate attention errors arising from KV cache quantization. KVLinC integrates two complementary techniques to enable robust low-precision caching. First, through a detailed analysis of Hadamard rotation based quantization strategies, we showed that applying channel-wise quantization to raw keys and token-wise quantization to Hadamard-transformed values minimizes quantization error. Second, to address residual errors from quantized keys, we proposed lightweight linear correction adapters that explicitly learn to compensate for distortions in attention. Extensive evaluation across the Llama, Qwen2.5, and Qwen3 model families demonstrates that KVLinC consistently matches or surpasses strong baselines under aggressive KV-cache compression. Finally, we developed a custom attention kernel that delivers up to $2.55\times$ speedup over FlashAttention, enabling scalable, efficient, and long-context LLM inference.

\subsubsection*{Reproducibility Statement}
We have provided details about our proposed algorithm in Section~\ref{sec:setup}. Additionally, we provide codebase to reproduce results of our experiments and the baselines in supplementary materials. 
\subsubsection*{Acknowledgments}
The authors would like to thank Sakshi Choudhary and Manish Nagaraj for helpful discussions. This work was supported by the Center for the Co-Design of Cognitive Systems (COCOSYS), a DARPA sponsored JUMP center of Semiconductor Research Corporation (SRC), Intel, SRC AIHW Program.

\bibliography{iclr2026_conference}

\begin{thebibliography}{43}
\providecommand{\natexlab}[1]{#1}
\providecommand{\url}[1]{\texttt{#1}}
\expandafter\ifx\csname urlstyle\endcsname\relax
  \providecommand{\doi}[1]{doi: #1}\else
  \providecommand{\doi}{doi: \begingroup \urlstyle{rm}\Url}\fi

\bibitem[Arora et~al.(2025)Arora, Eyuboglu, Zhang, Timalsina, Alberti, Zinsley, Zou, Rudra, and Ré]{arora2025based}
Simran Arora, Sabri Eyuboglu, Michael Zhang, Aman Timalsina, Silas Alberti, Dylan Zinsley, James Zou, Atri Rudra, and Christopher Ré.
\newblock Simple linear attention language models balance the recall-throughput tradeoff, 2025.
\newblock URL \url{https://arxiv.org/abs/2402.18668}.

\bibitem[Ashkboos et~al.(2024)Ashkboos, Mohtashami, Croci, Li, Cameron, Jaggi, Alistarh, Hoefler, and Hensman]{ashkboos2024quarot}
Saleh Ashkboos, Amirkeivan Mohtashami, Maximilian~L Croci, Bo~Li, Pashmina Cameron, Martin Jaggi, Dan Alistarh, Torsten Hoefler, and James Hensman.
\newblock Quarot: Outlier-free 4-bit inference in rotated llms.
\newblock \emph{Advances in Neural Information Processing Systems}, 37:\penalty0 100213--100240, 2024.

\bibitem[Bai et~al.(2023)Bai, Lv, Zhang, Lyu, Tang, Huang, Du, Liu, Zeng, Hou, et~al.]{bai2023longbench}
Yushi Bai, Xin Lv, Jiajie Zhang, Hongchang Lyu, Jiankai Tang, Zhidian Huang, Zhengxiao Du, Xiao Liu, Aohan Zeng, Lei Hou, et~al.
\newblock Longbench: A bilingual, multitask benchmark for long context understanding.
\newblock \emph{arXiv preprint arXiv:2308.14508}, 2023.

\bibitem[Chang et~al.(2024)Chang, Lin, Lin, Chen, Hu, Wang, Huang, Ceze, Abdelfattah, and Wu]{chang2024palu}
Chi-Chih Chang, Wei-Cheng Lin, Chien-Yu Lin, Chong-Yan Chen, Yu-Fang Hu, Pei-Shuo Wang, Ning-Chi Huang, Luis Ceze, Mohamed~S Abdelfattah, and Kai-Chiang Wu.
\newblock Palu: Compressing kv-cache with low-rank projection.
\newblock \emph{arXiv preprint arXiv:2407.21118}, 2024.

\bibitem[Choromanski et~al.(2020)Choromanski, Likhosherstov, Dohan, Song, Gane, Sarlos, Hawkins, Davis, Mohiuddin, Kaiser, et~al.]{choromanski2020performer}
Krzysztof Choromanski, Valerii Likhosherstov, David Dohan, Xingyou Song, Andreea Gane, Tamas Sarlos, Peter Hawkins, Jared Davis, Afroz Mohiuddin, Lukasz Kaiser, et~al.
\newblock Rethinking attention with performers.
\newblock \emph{arXiv preprint arXiv:2009.14794}, 2020.

\bibitem[Cobbe et~al.(2021)Cobbe, Kosaraju, Bavarian, Chen, Jun, Kaiser, Plappert, Tworek, Hilton, Nakano, et~al.]{cobbe2021gsm8k}
Karl Cobbe, Vineet Kosaraju, Mohammad Bavarian, Mark Chen, Heewoo Jun, Lukasz Kaiser, Matthias Plappert, Jerry Tworek, Jacob Hilton, Reiichiro Nakano, et~al.
\newblock Training verifiers to solve math word problems.
\newblock \emph{arXiv preprint arXiv:2110.14168}, 2021.

\bibitem[Dao(2023)]{dao2023flashattention}
Tri Dao.
\newblock Flashattention-2: Faster attention with better parallelism and work partitioning.
\newblock \emph{arXiv preprint arXiv:2307.08691}, 2023.

\bibitem[Dong et~al.(2024)Dong, Cheng, Qin, and Wang]{dong2024qaq}
Shichen Dong, Wen Cheng, Jiayu Qin, and Wei Wang.
\newblock Qaq: Quality adaptive quantization for llm kv cache.
\newblock \emph{arXiv preprint arXiv:2403.04643}, 2024.

\bibitem[Duanmu et~al.(2024)Duanmu, Yuan, Li, Duan, Zhang, and Lin]{duanmu2024skvq}
Haojie Duanmu, Zhihang Yuan, Xiuhong Li, Jiangfei Duan, Xingcheng Zhang, and Dahua Lin.
\newblock Skvq: Sliding-window key and value cache quantization for large language models.
\newblock \emph{arXiv preprint arXiv:2405.06219}, 2024.

\bibitem[Gao et~al.(2024)Gao, Tow, Abbasi, Biderman, Black, DiPofi, Foster, Golding, Hsu, Le~Noac'h, Li, McDonell, Muennighoff, Ociepa, Phang, Reynolds, Schoelkopf, Skowron, Sutawika, Tang, Thite, Wang, Wang, and Zou]{lm-eval-harness}
Leo Gao, Jonathan Tow, Baber Abbasi, Stella Biderman, Sid Black, Anthony DiPofi, Charles Foster, Laurence Golding, Jeffrey Hsu, Alain Le~Noac'h, Haonan Li, Kyle McDonell, Niklas Muennighoff, Chris Ociepa, Jason Phang, Laria Reynolds, Hailey Schoelkopf, Aviya Skowron, Lintang Sutawika, Eric Tang, Anish Thite, Ben Wang, Kevin Wang, and Andy Zou.
\newblock A framework for few-shot language model evaluation, 07 2024.
\newblock URL \url{https://zenodo.org/records/12608602}.

\bibitem[Han et~al.(2025)Han, Kacham, Karbasi, Mirrokni, and Zandieh]{han2025polarquant}
Insu Han, Praneeth Kacham, Amin Karbasi, Vahab Mirrokni, and Amir Zandieh.
\newblock Polarquant: Quantizing kv caches with polar transformation.
\newblock \emph{arXiv preprint arXiv:2502.02617}, 2025.

\bibitem[He et~al.(2024)He, Zhang, Wu, Liu, Zhou, and Zhuang]{he2024zipcache}
Yefei He, Luoming Zhang, Weijia Wu, Jing Liu, Hong Zhou, and Bohan Zhuang.
\newblock Zipcache: Accurate and efficient kv cache quantization with salient token identification.
\newblock \emph{Advances in Neural Information Processing Systems}, 37:\penalty0 68287--68307, 2024.

\bibitem[Hooper et~al.(2024)Hooper, Kim, Mohammadzadeh, Mahoney, Shao, Keutzer, and Gholami]{hooper2024kvquant}
Coleman Hooper, Sehoon Kim, Hiva Mohammadzadeh, Michael~W Mahoney, Yakun~S Shao, Kurt Keutzer, and Amir Gholami.
\newblock Kvquant: Towards 10 million context length llm inference with kv cache quantization.
\newblock \emph{Advances in Neural Information Processing Systems}, 37:\penalty0 1270--1303, 2024.

\bibitem[Hsieh et~al.(2024)Hsieh, Sun, Kriman, Acharya, Rekesh, Jia, Zhang, and Ginsburg]{hsieh2024ruler}
Cheng-Ping Hsieh, Simeng Sun, Samuel Kriman, Shantanu Acharya, Dima Rekesh, Fei Jia, Yang Zhang, and Boris Ginsburg.
\newblock Ruler: What's the real context size of your long-context language models?
\newblock \emph{arXiv preprint arXiv:2404.06654}, 2024.

\bibitem[Kang et~al.(2024)Kang, Zhang, Kundu, Jeong, Liu, Krishna, and Zhao]{kang2024gearefficientkvcache}
Hao Kang, Qingru Zhang, Souvik Kundu, Geonhwa Jeong, Zaoxing Liu, Tushar Krishna, and Tuo Zhao.
\newblock Gear: An efficient kv cache compression recipe for near-lossless generative inference of llm, 2024.
\newblock URL \url{https://arxiv.org/abs/2403.05527}.

\bibitem[Katharopoulos et~al.(2020)Katharopoulos, Vyas, Pappas, and Fleuret]{katharopoulos2020transformersarernns}
Angelos Katharopoulos, Apoorv Vyas, Nikolaos Pappas, and Fran{\c{c}}ois Fleuret.
\newblock Transformers are rnns: Fast autoregressive transformers with linear attention.
\newblock In \emph{International conference on machine learning}, pp.\  5156--5165. PMLR, 2020.

\bibitem[Kingma \& Ba(2017)Kingma and Ba]{kingma2017adammethodstochasticoptimization}
Diederik~P. Kingma and Jimmy Ba.
\newblock Adam: A method for stochastic optimization, 2017.
\newblock URL \url{https://arxiv.org/abs/1412.6980}.

\bibitem[Lan et~al.(2025)Lan, Sun, Hu, Du, and Cheng]{lan2025liger}
Disen Lan, Weigao Sun, Jiaxi Hu, Jusen Du, and Yu~Cheng.
\newblock Liger: Linearizing large language models to gated recurrent structures.
\newblock \emph{arXiv preprint arXiv:2503.01496}, 2025.

\bibitem[Liu et~al.(2024{\natexlab{a}})Liu, Zhao, Fedorov, Soran, Choudhary, Krishnamoorthi, Chandra, Tian, and Blankevoort]{liu2024spinquant}
Zechun Liu, Changsheng Zhao, Igor Fedorov, Bilge Soran, Dhruv Choudhary, Raghuraman Krishnamoorthi, Vikas Chandra, Yuandong Tian, and Tijmen Blankevoort.
\newblock Spinquant: Llm quantization with learned rotations.
\newblock \emph{arXiv preprint arXiv:2405.16406}, 2024{\natexlab{a}}.

\bibitem[Liu et~al.(2024{\natexlab{b}})Liu, Yuan, Jin, Zhong, Xu, Braverman, Chen, and Hu]{liu2024kivi}
Zirui Liu, Jiayi Yuan, Hongye Jin, Shaochen Zhong, Zhaozhuo Xu, Vladimir Braverman, Beidi Chen, and Xia Hu.
\newblock Kivi: A tuning-free asymmetric 2bit quantization for kv cache.
\newblock \emph{arXiv preprint arXiv:2402.02750}, 2024{\natexlab{b}}.

\bibitem[McDermott et~al.(2025)McDermott, Jr., and Parhi]{mcdermott2025lolalowranklinearattention}
Luke McDermott, Robert W.~Heath Jr., and Rahul Parhi.
\newblock Lola: Low-rank linear attention with sparse caching, 2025.
\newblock URL \url{https://arxiv.org/abs/2505.23666}.

\bibitem[Merity et~al.(2016)Merity, Xiong, Bradbury, and Socher]{merity2016wikitext}
Stephen Merity, Caiming Xiong, James Bradbury, and Richard Socher.
\newblock Pointer sentinel mixture models, 2016.
\newblock URL \url{https://arxiv.org/abs/1609.07843}.

\bibitem[Meta(2024{\natexlab{a}})]{llama3.2meta}
Meta.
\newblock {Llama 3.2: Revolutionizing edge AI and vision with open, customizable models}, 2024{\natexlab{a}}.
\newblock URL \url{https://ai.meta.com/blog/llama-3-2-connect-2024-vision-edge-mobile-devices/}.

\bibitem[Meta(2024{\natexlab{b}})]{llama3meta}
Meta.
\newblock {Introducing Meta Llama 3: The most capable openly available LLM to date.}, 2024{\natexlab{b}}.
\newblock URL \url{https://ai.meta.com/blog/meta-llama-3/}.

\bibitem[Paszke et~al.(2019)Paszke, Gross, Massa, Lerer, Bradbury, Chanan, Killeen, Lin, Gimelshein, Antiga, et~al.]{pytorch-Neurips2019}
Adam Paszke, Sam Gross, Francisco Massa, Adam Lerer, James Bradbury, Gregory Chanan, Trevor Killeen, Zeming Lin, Natalia Gimelshein, Luca Antiga, et~al.
\newblock Pytorch: An imperative style, high-performance deep learning library.
\newblock \emph{Advances in neural information processing systems}, 32, 2019.

\bibitem[Peters et~al.(2023)Peters, Fournarakis, Nagel, Van~Baalen, and Blankevoort]{peters2023qbitopt}
Jorn Peters, Marios Fournarakis, Markus Nagel, Mart Van~Baalen, and Tijmen Blankevoort.
\newblock Qbitopt: Fast and accurate bitwidth reallocation during training.
\newblock In \emph{Proceedings of the IEEE/CVF international conference on computer vision}, pp.\  1282--1291, 2023.

\bibitem[Saxena et~al.(2024{\natexlab{a}})Saxena, Saha, Choudhary, and Roy]{saxena2024eigen}
Utkarsh Saxena, Gobinda Saha, Sakshi Choudhary, and Kaushik Roy.
\newblock Eigen attention: Attention in low-rank space for kv cache compression.
\newblock \emph{arXiv preprint arXiv:2408.05646}, 2024{\natexlab{a}}.

\bibitem[Saxena et~al.(2024{\natexlab{b}})Saxena, Sharify, Roy, and Wang]{saxena2024resq}
Utkarsh Saxena, Sayeh Sharify, Kaushik Roy, and Xin Wang.
\newblock Resq: Mixed-precision quantization of large language models with low-rank residuals.
\newblock \emph{arXiv preprint arXiv:2412.14363}, 2024{\natexlab{b}}.

\bibitem[Suzgun et~al.(2022)Suzgun, Scales, Sch{\"a}rli, Gehrmann, Tay, Chung, Chowdhery, Le, Chi, Zhou, et~al.]{suzgun2022bigbenchhard}
Mirac Suzgun, Nathan Scales, Nathanael Sch{\"a}rli, Sebastian Gehrmann, Yi~Tay, Hyung~Won Chung, Aakanksha Chowdhery, Quoc~V Le, Ed~H Chi, Denny Zhou, et~al.
\newblock Challenging big-bench tasks and whether chain-of-thought can solve them.
\newblock \emph{arXiv preprint arXiv:2210.09261}, 2022.

\bibitem[Taori et~al.(2023)Taori, Gulrajani, Zhang, Dubois, Li, Guestrin, Liang, and Hashimoto]{taori2023alpaca}
Rohan Taori, Ishaan Gulrajani, Tianyi Zhang, Yann Dubois, Xuechen Li, Carlos Guestrin, Percy Liang, and Tatsunori~B Hashimoto.
\newblock Alpaca: A strong, replicable instruction-following model.
\newblock \emph{Stanford Center for Research on Foundation Models. https://crfm. stanford. edu/2023/03/13/alpaca. html}, 3\penalty0 (6):\penalty0 7, 2023.

\bibitem[Tillet et~al.(2019)Tillet, Kung, and Cox]{tillet2019triton}
Philippe Tillet, Hsiang-Tsung Kung, and David Cox.
\newblock Triton: an intermediate language and compiler for tiled neural network computations.
\newblock In \emph{Proceedings of the 3rd ACM SIGPLAN International Workshop on Machine Learning and Programming Languages}, pp.\  10--19, 2019.

\bibitem[Touvron et~al.(2023)Touvron, Martin, Stone, Albert, Almahairi, Babaei, Bashlykov, Batra, Bhargava, Bhosale, et~al.]{touvron2023llama2}
Hugo Touvron, Louis Martin, Kevin Stone, Peter Albert, Amjad Almahairi, Yasmine Babaei, Nikolay Bashlykov, Soumya Batra, Prajjwal Bhargava, Shruti Bhosale, et~al.
\newblock Llama 2: Open foundation and fine-tuned chat models.
\newblock \emph{arXiv preprint arXiv:2307.09288}, 2023.

\bibitem[Weber et~al.(2024)Weber, Fu, Anthony, Oren, Adams, Alexandrov, Lyu, Nguyen, Yao, Adams, et~al.]{weber2024redpajama}
Maurice Weber, Dan Fu, Quentin Anthony, Yonatan Oren, Shane Adams, Anton Alexandrov, Xiaozhong Lyu, Huu Nguyen, Xiaozhe Yao, Virginia Adams, et~al.
\newblock Redpajama: an open dataset for training large language models.
\newblock \emph{Advances in neural information processing systems}, 37:\penalty0 116462--116492, 2024.

\bibitem[Wolf et~al.(2020)Wolf, Debut, Sanh, Chaumond, Delangue, Moi, Cistac, Rault, Louf, Funtowicz, Davison, Shleifer, von Platen, Ma, Jernite, Plu, Xu, Scao, Gugger, Drame, Lhoest, and Rush]{huggingface-arxiv2019}
Thomas Wolf, Lysandre Debut, Victor Sanh, Julien Chaumond, Clement Delangue, Anthony Moi, Pierric Cistac, Tim Rault, Rémi Louf, Morgan Funtowicz, Joe Davison, Sam Shleifer, Patrick von Platen, Clara Ma, Yacine Jernite, Julien Plu, Canwen Xu, Teven~Le Scao, Sylvain Gugger, Mariama Drame, Quentin Lhoest, and Alexander~M. Rush.
\newblock Huggingface's transformers: State-of-the-art natural language processing.
\newblock \emph{arXiv:1910.03771}, 2020.

\bibitem[Xiong et~al.(2021)Xiong, Zeng, Chakraborty, Tan, Fung, Li, and Singh]{xiong2021nystromformer}
Yunyang Xiong, Zhanpeng Zeng, Rudrasis Chakraborty, Mingxing Tan, Glenn Fung, Yin Li, and Vikas Singh.
\newblock Nystr{\"o}mformer: A nystr{\"o}m-based algorithm for approximating self-attention.
\newblock In \emph{Proceedings of the AAAI conference on artificial intelligence}, volume~35, pp.\  14138--14148, 2021.

\bibitem[Yang et~al.(2024{\natexlab{a}})Yang, Yang, Zhang, Hui, Zheng, Yu, Li, Liu, Huang, Wei, et~al.]{yang2024qwen2}
An~Yang, Baosong Yang, Beichen Zhang, Binyuan Hui, Bo~Zheng, Bowen Yu, Chengyuan Li, Dayiheng Liu, Fei Huang, Haoran Wei, et~al.
\newblock {Qwen2.5} technical report.
\newblock \emph{arXiv:2412.15115}, 2024{\natexlab{a}}.

\bibitem[Yang et~al.(2025)Yang, Li, Yang, Zhang, Hui, Zheng, Yu, Gao, Huang, Lv, Zheng, Liu, Zhou, Huang, Hu, Ge, Wei, Lin, Tang, Yang, Tu, Zhang, Yang, Yang, Zhou, Zhou, Lin, Dang, Bao, Yang, Yu, Deng, Li, Xue, Li, Zhang, Wang, Zhu, Men, Gao, Liu, Luo, Li, Tang, Yin, Ren, Wang, Zhang, Ren, Fan, Su, Zhang, Zhang, Wan, Liu, Wang, Cui, Zhang, Zhou, and Qiu]{yang2025qwen3technicalreport}
An~Yang, Anfeng Li, Baosong Yang, Beichen Zhang, Binyuan Hui, Bo~Zheng, Bowen Yu, Chang Gao, Chengen Huang, Chenxu Lv, Chujie Zheng, Dayiheng Liu, Fan Zhou, Fei Huang, Feng Hu, Hao Ge, Haoran Wei, Huan Lin, Jialong Tang, Jian Yang, Jianhong Tu, Jianwei Zhang, Jianxin Yang, Jiaxi Yang, Jing Zhou, Jingren Zhou, Junyang Lin, Kai Dang, Keqin Bao, Kexin Yang, Le~Yu, Lianghao Deng, Mei Li, Mingfeng Xue, Mingze Li, Pei Zhang, Peng Wang, Qin Zhu, Rui Men, Ruize Gao, Shixuan Liu, Shuang Luo, Tianhao Li, Tianyi Tang, Wenbiao Yin, Xingzhang Ren, Xinyu Wang, Xinyu Zhang, Xuancheng Ren, Yang Fan, Yang Su, Yichang Zhang, Yinger Zhang, Yu~Wan, Yuqiong Liu, Zekun Wang, Zeyu Cui, Zhenru Zhang, Zhipeng Zhou, and Zihan Qiu.
\newblock Qwen3 technical report, 2025.
\newblock URL \url{https://arxiv.org/abs/2505.09388}.

\bibitem[Yang et~al.(2024{\natexlab{b}})Yang, Kim, Bae, Kwon, Park, Yang, Kwon, and Lee]{yang2024mikv}
June~Yong Yang, Byeongwook Kim, Jeongin Bae, Beomseok Kwon, Gunho Park, Eunho Yang, Se~Jung Kwon, and Dongsoo Lee.
\newblock No token left behind: Reliable kv cache compression via importance-aware mixed precision quantization.
\newblock \emph{arXiv preprint arXiv:2402.18096}, 2024{\natexlab{b}}.

\bibitem[Yang et~al.(2023)Yang, Wang, Shen, Panda, and Kim]{yang2023gatedlinearattention}
Songlin Yang, Bailin Wang, Yikang Shen, Rameswar Panda, and Yoon Kim.
\newblock Gated linear attention transformers with hardware-efficient training.
\newblock \emph{arXiv preprint arXiv:2312.06635}, 2023.

\bibitem[Yue et~al.(2024)Yue, Yuan, Duanmu, Zhou, Wu, and Nie]{yue2024wkvquant}
Yuxuan Yue, Zhihang Yuan, Haojie Duanmu, Sifan Zhou, Jianlong Wu, and Liqiang Nie.
\newblock Wkvquant: Quantizing weight and key/value cache for large language models gains more.
\newblock \emph{arXiv preprint arXiv:2402.12065}, 2024.

\bibitem[Zandieh et~al.(2025)Zandieh, Daliri, and Han]{zandieh2025qjl}
Amir Zandieh, Majid Daliri, and Insu Han.
\newblock Qjl: 1-bit quantized jl transform for kv cache quantization with zero overhead.
\newblock In \emph{Proceedings of the AAAI Conference on Artificial Intelligence}, volume~39, pp.\  25805--25813, 2025.

\bibitem[Zhang et~al.(2024)Zhang, Arora, Chalamala, Wu, Spector, Singhal, Ramesh, and R{\'e}]{zhang2024lolcats}
Michael Zhang, Simran Arora, Rahul Chalamala, Alan Wu, Benjamin Spector, Aaryan Singhal, Krithik Ramesh, and Christopher R{\'e}.
\newblock Lolcats: On low-rank linearizing of large language models.
\newblock \emph{arXiv preprint arXiv:2410.10254}, 2024.

\bibitem[Zhou et~al.(2023)Zhou, Lu, Mishra, Brahma, Basu, Luan, Zhou, and Hou]{zhou2023ifeval}
Jeffrey Zhou, Tianjian Lu, Swaroop Mishra, Siddhartha Brahma, Sujoy Basu, Yi~Luan, Denny Zhou, and Le~Hou.
\newblock Instruction-following evaluation for large language models.
\newblock \emph{arXiv preprint arXiv:2311.07911}, 2023.

\end{thebibliography}
\bibliographystyle{iclr2026_conference}

\appendix
\section{Appendix}
\subsection{Detailed Results on LongBench tasks} \label{sec:app_longbench}
Here we show task-wise accuracy on various tasks within the LongBench benchmark \cite{bai2023longbench}. The results are presented in Table~\ref{tab:appendix_results_longbench}. We evaluate on 14 english language tasks on LongBench. Both KIVI and KVLinC show comparable performance on various tasks. 
\begin{table}[h!]
\centering
\caption{Taskwise accuracy on LongBench tasks.\textsuperscript{*}Upper bound performance.}
\label{tab:appendix_results_longbench}
\resizebox{\textwidth}{!}{
\begin{tabular}{c|c|c|ccccccccccccccc}
\hline
 &  &  & \multicolumn{15}{c}{LongBench Tasks} \\ \cline{4-18} 
\multirow{-2}{*}{Model} & \multirow{-2}{*}{Method} & \multirow{-2}{*}{\begin{tabular}[c]{@{}c@{}}KV \\ Cache\end{tabular}} & \begin{tabular}[c]{@{}c@{}}Multi\\  News\end{tabular} & \begin{tabular}[c]{@{}c@{}}Passage \\ Count\end{tabular} & Samsum & MFQA & \begin{tabular}[c]{@{}c@{}}Narrative\\ QA\end{tabular} & \begin{tabular}[c]{@{}c@{}}Hotpot\\ QA\end{tabular} & Trec & Qmsum & \begin{tabular}[c]{@{}c@{}}Trivia\\ QA\end{tabular} & Qasper & \begin{tabular}[c]{@{}c@{}}2Wiki\\ Mqa\end{tabular} & Musique & \begin{tabular}[c]{@{}c@{}}Gov \\ Report\end{tabular} & \multicolumn{1}{c|}{\begin{tabular}[c]{@{}c@{}}Passage\\  Retrieval\end{tabular}} & Avg. \\ \hline
 & FP16 & 16 & 26.2 & 3.5 & 42.5 & 51.1 & 26.3 & 30.3 & 71.0 & 22.7 & 88.9 & 40.6 & 28.0 & 13.7 & 33.5 & \multicolumn{1}{c|}{86.8} & 40.4 \\
 & KIVI & 2.46 & 24.7 & \textbf{3.6} & 40.7 & \textbf{48.5} & \textbf{27.1} & \textbf{30.6} & \textbf{70.5} & 23.9 & \textbf{88.7} & 36.1 & \textbf{32.5} & 13.9 & 26.4 & \multicolumn{1}{c|}{83.7} & \textbf{39.4} \\
\multirow{-3}{*}{\begin{tabular}[c]{@{}c@{}}Llama-\\ 3.2-3B\\ -Instruct\end{tabular}} & \cellcolor[HTML]{CEEFF8}KVLinC & \cellcolor[HTML]{CEEFF8}2.71 & \cellcolor[HTML]{CEEFF8}\textbf{26.2} & \cellcolor[HTML]{CEEFF8}2.5 & \cellcolor[HTML]{CEEFF8}\textbf{41.7} & \cellcolor[HTML]{CEEFF8}47.6 & \cellcolor[HTML]{CEEFF8}26.7 & \cellcolor[HTML]{CEEFF8}29.0 & \cellcolor[HTML]{CEEFF8}\textbf{70.5} & \cellcolor[HTML]{CEEFF8}\textbf{24.2} & \cellcolor[HTML]{CEEFF8}87.6 & \cellcolor[HTML]{CEEFF8}\textbf{36.3} & \cellcolor[HTML]{CEEFF8}31.4 & \cellcolor[HTML]{CEEFF8}\textbf{14.4} & \cellcolor[HTML]{CEEFF8}\textbf{29.9} & \multicolumn{1}{c|}{\cellcolor[HTML]{CEEFF8}\textbf{84.3}} & \cellcolor[HTML]{CEEFF8}\textbf{39.4} \\ \hline
 & FP16 & 16 & 24.8 & 3.0 & 44.2 & 38.7 & 10.9 & 19.9 & 68.5 & 23.4 & 87.1 & 16.4 & 15.2 & 12.4 & 32.4 & \multicolumn{1}{c|}{42.8} & 31.4 \\
 & KIVI & 2.46 & \textbf{23.5} & \textbf{3.0} & \textbf{42.2} & 28.1 & 9.2 & \textbf{18.3} & \textbf{68.0} & \textbf{24.3} & 85.6 & 11.4 & \textbf{13.2} & \textbf{9.7} & 24.3 & \multicolumn{1}{c|}{31.2} & 28.0 \\
\multirow{-3}{*}{\begin{tabular}[c]{@{}c@{}}Qwen-\\ 2.5-3B-\\ Instruct\end{tabular}} & \cellcolor[HTML]{CEEFF8}KVLinC & \cellcolor[HTML]{CEEFF8}2.71 & \cellcolor[HTML]{CEEFF8}23.0 & \cellcolor[HTML]{CEEFF8}2.2 & \cellcolor[HTML]{CEEFF8}41.6 & \cellcolor[HTML]{CEEFF8}\textbf{31.0} & \cellcolor[HTML]{CEEFF8}\textbf{10.1} & \cellcolor[HTML]{CEEFF8}14.2 & \cellcolor[HTML]{CEEFF8}\textbf{68.0} & \cellcolor[HTML]{CEEFF8}24.1 & \cellcolor[HTML]{CEEFF8}\textbf{86.6} & \cellcolor[HTML]{CEEFF8}\textbf{12.1} & \cellcolor[HTML]{CEEFF8}\textbf{13.2} & \cellcolor[HTML]{CEEFF8}9.0 & \cellcolor[HTML]{CEEFF8}\textbf{27.3} & \multicolumn{1}{c|}{\cellcolor[HTML]{CEEFF8}\textbf{31.8}} & \cellcolor[HTML]{CEEFF8}\textbf{28.2} \\ \hline
 & FP16 & 16 & 19.8 & 3.3 & 44.1 & 24.8 & 3.5 & 13.4 & 73.0 & 23.4 & 88.8 & 11.1 & 14.4 & 10.0 & 29.2 & \multicolumn{1}{c|}{88.0} & 31.9 \\
 & KIVI & 2.46 & \textbf{22.9} & \textbf{4.6} & \textbf{42.0} & 21.3 & 3.7 & \textbf{13.1} & \textbf{73.0} & \textbf{22.4} & 87.8 & 10.7 & \textbf{13.6} & 7.9 & 26.2 & \multicolumn{1}{c|}{\textbf{87.5}} & 31.2 \\
\multirow{-3}{*}{\begin{tabular}[c]{@{}c@{}}Qwen\\ -3-4B-\\ Instruct\end{tabular}} & \cellcolor[HTML]{CEEFF8}KVLinC & \cellcolor[HTML]{CEEFF8}2.71 & \cellcolor[HTML]{CEEFF8}22.7 & \cellcolor[HTML]{CEEFF8}3.2 & \cellcolor[HTML]{CEEFF8}41.9 & \cellcolor[HTML]{CEEFF8}\textbf{21.8} & \cellcolor[HTML]{CEEFF8}\textbf{4.3} & \cellcolor[HTML]{CEEFF8}12.8 & \cellcolor[HTML]{CEEFF8}\textbf{73.0} & \cellcolor[HTML]{CEEFF8}23.2 & \cellcolor[HTML]{CEEFF8}\textbf{88.3} & \cellcolor[HTML]{CEEFF8}\textbf{11.0} & \cellcolor[HTML]{CEEFF8}12.9 & \cellcolor[HTML]{CEEFF8}\textbf{8.9} & \cellcolor[HTML]{CEEFF8}\textbf{27.1} & \multicolumn{1}{c|}{\cellcolor[HTML]{CEEFF8}83.6} & \cellcolor[HTML]{CEEFF8}31.0 \\ \hline
\end{tabular}
}
\end{table}
\subsection{Detailed Results on RULER tasks}
\begin{table}[t!]
\centering
\caption{Task-wise accuacy on RULER benchmark\textsuperscript{*}Upper bound performance.}
\label{tab:appendix_results_ruler}
\resizebox{\textwidth}{!}{
\begin{tabular}{c|c|c|c|cccccccccccccc}
\hline
 &  &  &  & \multicolumn{14}{c}{RULER Tasks} \\ \cline{5-18} 
\multirow{-2}{*}{Model} & \multirow{-2}{*}{Seq-len} & \multirow{-2}{*}{Method} & \multirow{-2}{*}{\begin{tabular}[c]{@{}c@{}}KV \\ Cache\end{tabular}} & niahm1 & niahm2 & niahm3 & \begin{tabular}[c]{@{}c@{}}niah\\ multiq\end{tabular} & \begin{tabular}[c]{@{}c@{}}niah\\ mltiV\end{tabular} & niahs1 & niahs2 & niahs3 & cwe & fwe & hotpotqa & squadqa & \multicolumn{1}{c|}{vt} & Avg. \\ \hline
 &  & FP16 & 16 & 99.8 & 100.0 & 98.4 & 100.0 & 99.8 & 100.0 & 100.0 & 99.6 & 95.1 & 93.1 & 55.2 & 68.9 & \multicolumn{1}{c|}{92.1} & 92.5 \\
 &  & KIVI & 2.46 & \textbf{98.8} & 89.0 & 20.2 & 91.6 & 90.6 & \textbf{99.0} & \textbf{99.6} & 53.0 & 79.2 & 86.5 & \textbf{52.0} & 66.4 & \multicolumn{1}{c|}{70.6} & 76.7 \\
 & \multirow{-3}{*}{4k} & \cellcolor[HTML]{CEEFF8}KVLinC & \cellcolor[HTML]{CEEFF8}2.71 & \cellcolor[HTML]{CEEFF8}95.4 & \cellcolor[HTML]{CEEFF8}\textbf{94.4} & \cellcolor[HTML]{CEEFF8}\textbf{23.4} & \cellcolor[HTML]{CEEFF8}\textbf{93.9} & \cellcolor[HTML]{CEEFF8}\textbf{95.3} & \cellcolor[HTML]{CEEFF8}96.0 & \cellcolor[HTML]{CEEFF8}98.8 & \cellcolor[HTML]{CEEFF8}\textbf{80.4} & \cellcolor[HTML]{CEEFF8}\textbf{85.8} & \cellcolor[HTML]{CEEFF8}\textbf{88.2} & \cellcolor[HTML]{CEEFF8}51.8 & \cellcolor[HTML]{CEEFF8}\textbf{69.2} & \multicolumn{1}{c|}{\cellcolor[HTML]{CEEFF8}\textbf{77.4}} & \cellcolor[HTML]{CEEFF8}\textbf{80.8} \\ \cline{2-18} 
 &  & FP16 & 16 & 98.4 & 99.8 & 96.0 & 99.5 & 99.5 & 100.0 & 100.0 & 99.8 & 66.9 & 85.6 & 52.6 & 63.8 & \multicolumn{1}{c|}{84.0} & 88.1 \\
 &  & KIVI & 2.46 & \textbf{97.0} & 82.8 & 5.6 & 90.1 & 92.0 & \textbf{99.4} & \textbf{99.0} & 55.2 & 38.9 & 74.3 & \textbf{51.2} & 56.3 & \multicolumn{1}{c|}{72.0} & 70.3 \\
\multirow{-6}{*}{\begin{tabular}[c]{@{}c@{}}Llama-\\ 3.2-3B\\ -Instruct\end{tabular}} & \multirow{-3}{*}{8k} & \cellcolor[HTML]{CEEFF8}KVLinC & \cellcolor[HTML]{CEEFF8}2.71 & \cellcolor[HTML]{CEEFF8}93.0 & \cellcolor[HTML]{CEEFF8}\textbf{92.6} & \cellcolor[HTML]{CEEFF8}\textbf{10.2} & \cellcolor[HTML]{CEEFF8}\textbf{92.7} & \cellcolor[HTML]{CEEFF8}\textbf{92.5} & \cellcolor[HTML]{CEEFF8}93.6 & \cellcolor[HTML]{CEEFF8}96.8 & \cellcolor[HTML]{CEEFF8}\textbf{73.2} & \cellcolor[HTML]{CEEFF8}\textbf{50.0} & \cellcolor[HTML]{CEEFF8}\textbf{80.9} & \cellcolor[HTML]{CEEFF8}49.4 & \cellcolor[HTML]{CEEFF8}\textbf{59.8} & \multicolumn{1}{c|}{\cellcolor[HTML]{CEEFF8}\textbf{72.4}} & \cellcolor[HTML]{CEEFF8}\textbf{73.6} \\ \hline
 &  & FP16 & 16 & 99.8 & 99.0 & 97.4 & 100.0 & 99.5 & 100.0 & 84.7 & 99.8 & 84.7 & 91.5 & 49.0 & 72.1 & \multicolumn{1}{c|}{96.6} & 90.3 \\
 &  & KIVI & 2.46 & 65.0 & 47.6 & 0.0 & 58.5 & 44.7 & 66.0 & 56.6 & 2.4 & \textbf{65.3} & 79.9 & \textbf{43.0} & 62.3 & \multicolumn{1}{c|}{51.9} & 49.5 \\
 & \multirow{-3}{*}{4k} & \cellcolor[HTML]{CEEFF8}KVLinC & \cellcolor[HTML]{CEEFF8}2.71 & \cellcolor[HTML]{CEEFF8}\textbf{82.4} & \cellcolor[HTML]{CEEFF8}\textbf{52.6} & \cellcolor[HTML]{CEEFF8}\textbf{0.6} & \cellcolor[HTML]{CEEFF8}\textbf{75.7} & \cellcolor[HTML]{CEEFF8}\textbf{70.4} & \cellcolor[HTML]{CEEFF8}\textbf{87.6} & \cellcolor[HTML]{CEEFF8}\textbf{87.6} & \cellcolor[HTML]{CEEFF8}\textbf{16.6} & \cellcolor[HTML]{CEEFF8}65.2 & \cellcolor[HTML]{CEEFF8}\textbf{81.7} & \cellcolor[HTML]{CEEFF8}42.8 & \cellcolor[HTML]{CEEFF8}\textbf{64.4} & \multicolumn{1}{c|}{\cellcolor[HTML]{CEEFF8}\textbf{63.8}} & \cellcolor[HTML]{CEEFF8}\textbf{60.9} \\ \cline{3-18} 
 &  & FP16 & 16 & 100.0 & 99.6 & 87.6 & 100.0 & 98.4 & 100.0 & 100.0 & 100.0 & 46.3 & 77.1 & 43.6 & 58.5 & \multicolumn{1}{c|}{94.5} & 85.0 \\
 &  & KIVI & 2.46 & 57.8 & 26.4 & \textbf{0.0} & 55.3 & 39.7 & 69.0 & 56.6 & 3.8 & \textbf{34.2} & 61.4 & 34.8 & 45.3 & \multicolumn{1}{c|}{49.6} & 41.1 \\
\multirow{-6}{*}{\begin{tabular}[c]{@{}c@{}}Qwen\\ -2.5-3B-\\ Instruct\end{tabular}} & \multirow{-3}{*}{8k} & \cellcolor[HTML]{CEEFF8}KVLinC & \cellcolor[HTML]{CEEFF8}2.71 & \cellcolor[HTML]{CEEFF8}\textbf{75.8} & \cellcolor[HTML]{CEEFF8}\textbf{34.6} & \cellcolor[HTML]{CEEFF8}\textbf{0.0} & \cellcolor[HTML]{CEEFF8}\textbf{69.9} & \cellcolor[HTML]{CEEFF8}\textbf{64.4} & \cellcolor[HTML]{CEEFF8}\textbf{85.6} & \cellcolor[HTML]{CEEFF8}\textbf{81.2} & \cellcolor[HTML]{CEEFF8}\textbf{21.6} & \cellcolor[HTML]{CEEFF8}29.6 & \cellcolor[HTML]{CEEFF8}\textbf{65.3} & \cellcolor[HTML]{CEEFF8}\textbf{36.4} & \cellcolor[HTML]{CEEFF8}\textbf{48.0} & \multicolumn{1}{c|}{\cellcolor[HTML]{CEEFF8}\textbf{52.4}} & \cellcolor[HTML]{CEEFF8}\textbf{51.1} \\ \hline
 &  & FP16 & 16 & 97.4 & 100.0 & 99.8 & 99.6 & 98.3 & 100.0 & 100.0 & 99.8 & 94.7 & 88.1 & 54.8 & 72.1 & \multicolumn{1}{c|}{100.0} & 92.7 \\
 &  & KIVI & 2.46 & 96.6 & 95.0 & 48.0 & 97.4 & 96.6 & \textbf{99.2} & 96.8 & 78.8 & \textbf{81.9} & 82.1 & 55.6 & 69.6 & \multicolumn{1}{c|}{90.9} & 83.7 \\
 & \multirow{-3}{*}{4k} & \cellcolor[HTML]{CEEFF8}KVLinC & \cellcolor[HTML]{CEEFF8}2.71 & \cellcolor[HTML]{CEEFF8}\textbf{97.4} & \cellcolor[HTML]{CEEFF8}\textbf{96.6} & \cellcolor[HTML]{CEEFF8}\textbf{63.6} & \cellcolor[HTML]{CEEFF8}\textbf{98.1} & \cellcolor[HTML]{CEEFF8}\textbf{98.0} & \cellcolor[HTML]{CEEFF8}99.0 & \cellcolor[HTML]{CEEFF8}\textbf{97.8} & \cellcolor[HTML]{CEEFF8}\textbf{84.6} & \cellcolor[HTML]{CEEFF8}81.2 & \cellcolor[HTML]{CEEFF8}\textbf{83.1} & \cellcolor[HTML]{CEEFF8}\textbf{56.0} & \cellcolor[HTML]{CEEFF8}\textbf{71.3} & \multicolumn{1}{c|}{\cellcolor[HTML]{CEEFF8}\textbf{94.4}} & \cellcolor[HTML]{CEEFF8}\textbf{86.2} \\ \cline{2-18} 
 &  & FP16 & 16 & 97.8 & 99.0 & 99.4 & 99.3 & 96.2 & 100.0 & 100.0 & 100.0 & 66.7 & 83.6 & 50.6 & 59.8 & \multicolumn{1}{c|}{99.2} & 88.6 \\
 &  & KIVI & 2.46 & 96.2 & 91.2 & 26.4 & 96.6 & 96.4 & 99.2 & 96.0 & 75.4 & 67.8 & \textbf{82.1} & 55.4 & \textbf{63.7} & \multicolumn{1}{c|}{92.2} & 79.9 \\
\multirow{-6}{*}{\begin{tabular}[c]{@{}c@{}}Qwen-\\ 3-4B-\\ Instruct\end{tabular}} & \multirow{-3}{*}{8k} & \cellcolor[HTML]{CEEFF8}KVLinC & \cellcolor[HTML]{CEEFF8}2.71 & \cellcolor[HTML]{CEEFF8}\textbf{97.8} & \cellcolor[HTML]{CEEFF8}\textbf{94.2} & \cellcolor[HTML]{CEEFF8}\textbf{42.2} & \cellcolor[HTML]{CEEFF8}\textbf{98.2} & \cellcolor[HTML]{CEEFF8}\textbf{97.7} & \cellcolor[HTML]{CEEFF8}\textbf{99.8} & \cellcolor[HTML]{CEEFF8}\textbf{96.6} & \cellcolor[HTML]{CEEFF8}\textbf{83.2} & \cellcolor[HTML]{CEEFF8}\textbf{69.5} & \cellcolor[HTML]{CEEFF8}79.8 & \cellcolor[HTML]{CEEFF8}\textbf{55.6} & \cellcolor[HTML]{CEEFF8}63.1 & \multicolumn{1}{c|}{\cellcolor[HTML]{CEEFF8}\textbf{93.1}} & \cellcolor[HTML]{CEEFF8}\textbf{82.4} \\ \hline
\end{tabular}
}
\end{table}
\label{sec:app_ruler}
Additionally we also provide task wise breakdown in RULER benchmark in Table~\ref{tab:appendix_results_ruler}. The results are presented for both 4k and 8k sequence length. As shown in Table~\ref{tab:appendix_results_ruler}, KVLinC outperforms KIVI on most of the individual tasks across sequence lengths and models. 
\subsection{LLM Usage}
The authors of this paper used ChatGPT (\url{https://chatgpt.com/}) for polishing text within this paper. The authors take full responsibility for the content within this paper.

\end{document}